%% file: main.tex
\documentclass[10pt,twocolumn,letterpaper]{article}

\usepackage[pagenumbers]{cvpr}

\input{preamble}
\usepackage{float}
\usepackage{stfloats}
\usepackage{xspace}
\usepackage{booktabs}
\usepackage{array}
\usepackage{multirow}
\usepackage{graphicx}
\usepackage{amsmath,amssymb}
\usepackage{xcolor}
\usepackage{enumitem}
\usepackage{pifont}
\usepackage{tcolorbox}
\usepackage[scaled=0.95]{helvet}
\usepackage{fontawesome5}
\definecolor{leaftitlebg}{HTML}{F0EDF7}
\definecolor{leaflink}{HTML}{72549A}
\newcommand{\bench}{LEAF\xspace}

\newcommand{\Aone}{\mathrm{A}_1}
\newcommand{\Athree}{\mathrm{A}_3}

\definecolor{cvprblue}{rgb}{0.21,0.49,0.74}
\usepackage[pagebackref,breaklinks,linktoc=all,bookmarksnumbered,colorlinks,allcolors=cvprblue]{hyperref}

\title{\bench: Do Agent-as-Policy Robots Notice and Fix Their Own Failures?}

\input{authors}

\begin{document}
\onecolumn
\input{frontpage}
\clearpage
\twocolumn[{%
\begin{minipage}{\textwidth}
  \captionsetup{type=figure}
  \centering
  \IfFileExists{figs/LEF_Bench.pdf}{\includegraphics[width=0.92\linewidth,trim=8bp 8bp 8bp 8bp,clip]{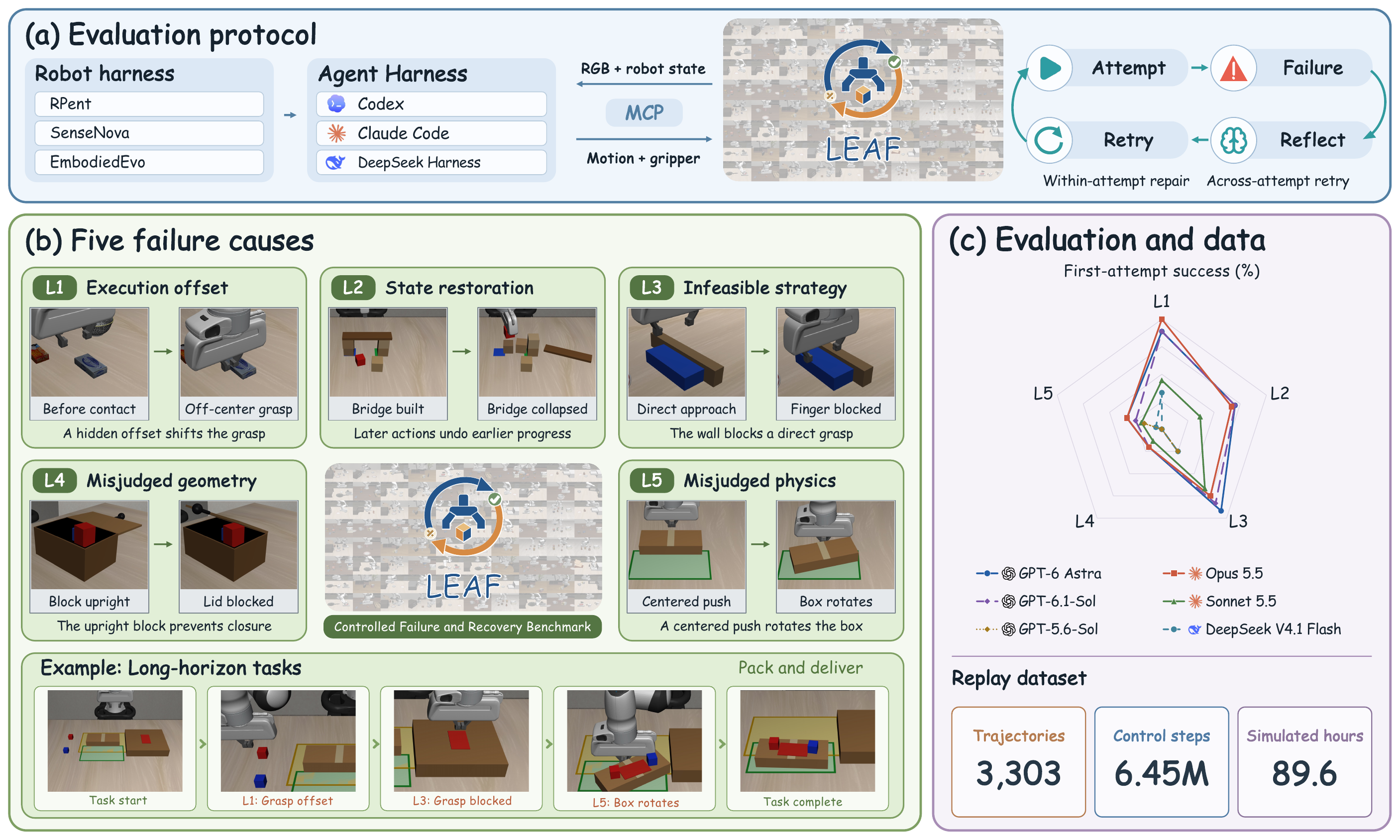}}{\fbox{\parbox{0.95\linewidth}{\centering\vspace{3em}main figure\vspace{3em}}}}
  \caption{\textbf{\bench at a glance.}
  (a)~Robot and agent harnesses are interchangeable above one MCP interface; an agent attempts, fails, reflects and retries, repairing within the attempt or retrying after a reset.
  (b)~The five causes of first-attempt failure, shown with frames, and a long-horizon task (L6).
  (c)~First-attempt success across five causes for six models, along with replay dataset statistics.}
  \label{fig:teaser}
\end{minipage}
\par\vspace{14pt}
}]
\input{sections/1_intro}
\input{sections/2_related}
\input{sections/3_benchmark}
\input{sections/4_experiments_v2}

\input{sections/5_conclusion}
\clearpage
\input{sections/6_llm_usage}
{
    \small
    \bibliographystyle{ieeenat_fullname}
    \bibliography{refs,refs_additions}
}

\clearpage
\input{supplementary}

\end{document}

%% file: preamble.tex
\usepackage{microtype}

\usepackage{needspace}
\newcommand{\finding}[2]{\paragraph{Finding~#1. #2}}

\makeatletter
\renewcommand{\paragraph}{\@startsection{paragraph}{4}{\z@}%
  {0.50\baselineskip}{-0.65em}{\normalfont\normalsize\bfseries}}
\makeatother

\usepackage{longtable,calc,ragged2e,fvextra,placeins}
\usepackage[nospread]{cuted}
\DeclareUnicodeCharacter{00D7}{\ensuremath{\times}}
\DeclareUnicodeCharacter{27E8}{\ensuremath{\langle}}
\DeclareUnicodeCharacter{27E9}{\ensuremath{\rangle}}
\providecommand{\tightlist}{\setlength{\itemsep}{0pt}\setlength{\parskip}{0pt}}

%% file: authors.tex
\newcommand{\leafauthors}{%
Yuxiang Xie\textsuperscript{1,2},
Chenyu Zhang\textsuperscript{3},
Ho-Jui Fang\textsuperscript{3},
LU Enqiao\textsuperscript{1},
Yu-Chiao Huang\textsuperscript{3},
Xuanting Ding\textsuperscript{3},
Jiaxin Fang\textsuperscript{2},
Junqi Jing\textsuperscript{2},
Pingyue Sheng\textsuperscript{2,3},
Weikang Wan\textsuperscript{2,4},
Yuxiang Gao\textsuperscript{2,\textdagger},
Mengdi Xu\textsuperscript{3,\textdagger}}
\newcommand{\leafaffiliations}{%
\textsuperscript{1}The Chinese University of Hong Kong, Shenzhen\quad
\textsuperscript{2}COCOMatrix\\[2pt]
\textsuperscript{3}Tsinghua University\quad
\textsuperscript{4}University of California, San Diego}
\newcommand{\leafauthornote}{%
\textsuperscript{\textdagger}Corresponding authors.}
\author{\leafauthors}

%% file: frontpage.tex
\thispagestyle{empty}
\begin{tcolorbox}[
  colback=leaftitlebg,colframe=leaftitlebg,boxrule=0pt,
  arc=10pt,left=17pt,right=17pt,top=18pt,bottom=18pt,
  before skip=0pt,after skip=0pt]
{\sffamily\bfseries\fontsize{22}{25}\selectfont\raggedright
LEAF: Do Agent-as-Policy Robots Notice and Fix Their Own Failures?\par}
\vspace{12pt}
{\sffamily\bfseries\fontsize{10}{13}\selectfont\raggedright
\leafauthors\par}
\vspace{7pt}
{\fontsize{10}{12}\selectfont\raggedright
\leafaffiliations\par}
\vspace{3pt}
{\fontsize{9}{11}\selectfont\leafauthornote\par}
\vspace{17pt}
\begingroup
\fontsize{11}{14}\selectfont
\setlength{\parindent}{0pt}
\renewenvironment{abstract}{\ignorespaces}{\par}
\input{sections/0_abstract}

\endgroup
\vspace{12pt}
{\fontsize{10}{13}\selectfont\urlstyle{same}\raggedright
\faEnvelope\enspace Email:\enspace
\href{mailto:124090742@link.cuhk.edu.cn}{\textcolor{leaflink}{\nolinkurl{124090742@link.cuhk.edu.cn}}}\\[2pt]
\faGlobe\enspace Project page:\enspace
\href{https://peter-cuhk.github.io/LEAF/}{\textcolor{leaflink}{\nolinkurl{https://peter-cuhk.github.io/LEAF/}}}\\[2pt]
\faDatabase\enspace Dataset:\enspace
\href{https://huggingface.co/datasets/COCOMatrix/LEAF-Bench}{\textcolor{leaflink}{\nolinkurl{huggingface.co/datasets/COCOMatrix/LEAF-Bench}}}\par}
\vspace{22pt}
\noindent\hfill
\raisebox{-0.5\height}{\includegraphics[height=31pt,trim=0bp 0bp 571bp 0bp,clip]{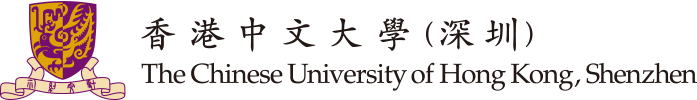}}\hspace{19pt}
\raisebox{-0.5\height}{\includegraphics[width=93pt]{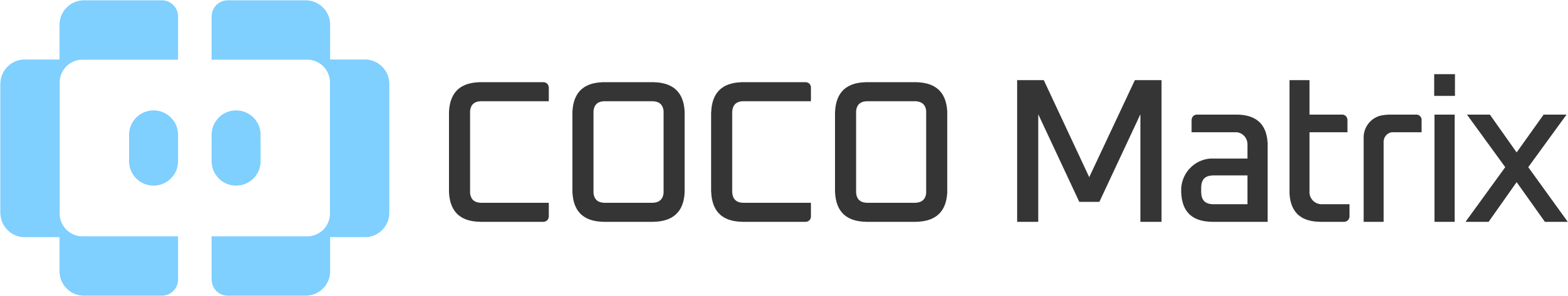}}\hspace{19pt}
\raisebox{-0.5\height}{\includegraphics[height=31pt]{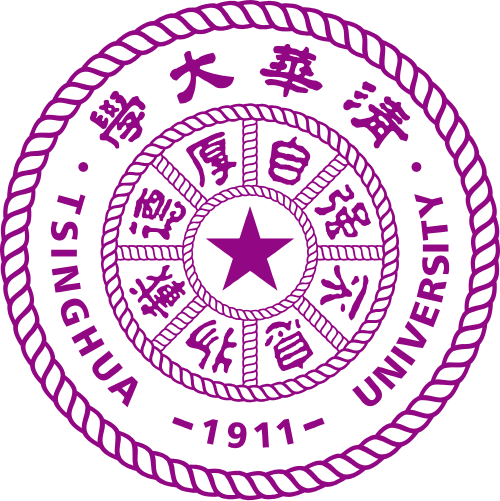}}\hspace{19pt}
\raisebox{-0.5\height}{\includegraphics[width=94pt]{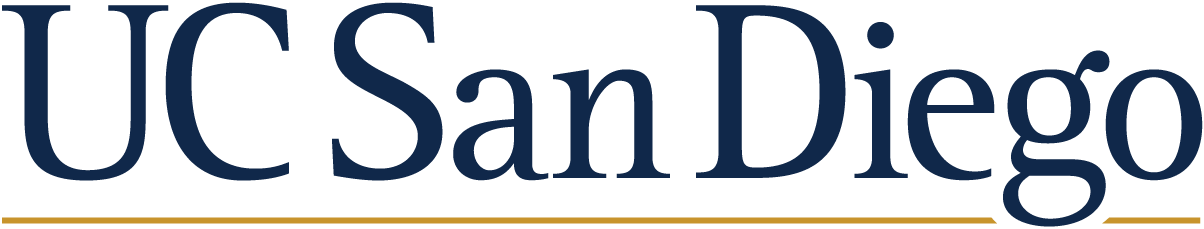}}
\end{tcolorbox}

%% file: sections/0_abstract.tex
\begin{abstract}
General-purpose multimodal agents increasingly serve as robot policies that perceive, reason, and act through tools, but whether they use their own failures to improve subsequent actions remains unclear. Existing evaluations conflate clean execution with recovery, while success over repeated attempts can improve through resampling alone. We introduce LEAF (\textbf{Lea}rning From \textbf{F}ailure Benchmark), comprising 40 manipulation tasks built around hidden traps that cannot be resolved by blindly repeating the initial strategy. LEAF supports interchangeable agent and robot harnesses above one MCP interface, while a fine-grained evaluator separates clean success, in-attempt repair, and cross-attempt recovery against memory-free controls. Evaluations establish causal diagnosis as the primary bottleneck for frontier agents: revealing the hidden cause largely eliminates their errors, while weaker agents struggle with diagnosis and execution. VLA and WAM policies show limited recovery after success-only fine-tuning. Retaining interaction history consistently outperforms memory-free retries, yet agents rarely restore undone subgoals and often rationalize subtle anomalies into false completion claims. LEAF-Lite, a prompt-only robot harness derived from these insights, achieves the highest observed first- and three-attempt success rates in our controlled harness comparison while reducing token use. We release the benchmark, code, interface, and over 6 million labeled replay frames with agent reasoning traces.
\end{abstract}

%% file: sections/1_intro.tex
\suppressfloats[t]
\section{Introduction}
\label{sec:intro}

General-purpose multimodal agents increasingly serve directly as robot policies, interpreting camera images and issuing control commands through code or tool interfaces~\cite{liang2023codeaspolicies,huang2023voxposer,fu2026capx,diao2026liberoagent,yang2026robotworld,liu2026roboquest}.
In real-world environments, policies frequently fail on their first try and must learn from failure through interaction.
Yet existing benchmarks across diverse tasks and embodiments~\cite{diao2026liberoagent,yang2026robotworld,liu2026roboquest} primarily report overall episode success, leaving a fundamental question unanswered: \emph{Can these agents detect their own failures, diagnose the causes, and adapt their actions?}

\begin{figure}[t]
  \centering
  \includegraphics[width=\columnwidth]{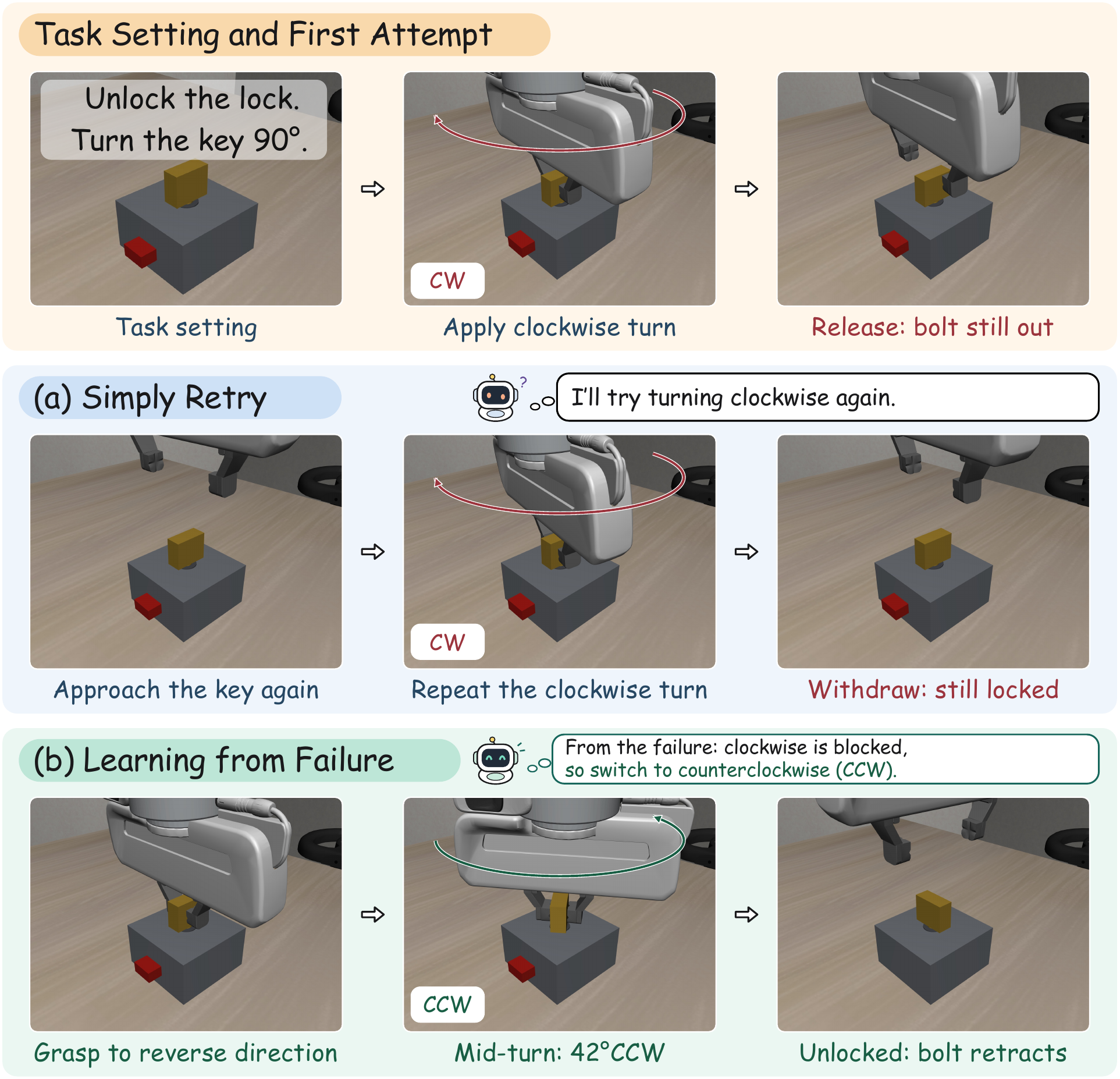}
  \caption{\textbf{Blind retry versus learning from failure.}
  A retry can resolve a \emph{slip}. The stalled clockwise turn is a \emph{trap}. The agent detects the blockage and turns counterclockwise to open the lock.}
  \label{fig:motivation}
\end{figure}

Execution failures in robotic manipulation fall into two categories: \emph{slips} and \emph{traps}.
\emph{Slips}, such as a missed grasp, stem from minor execution deviations during fine-grained manipulation and can typically be resolved by simply repeating the action.
In contrast, \emph{traps} originate from the policy itself when actions guided by training priors or common sense fail under specific task conditions.
In the key-turning task in \cref{fig:motivation}, for example, most policies initially attempt a standard clockwise turn that stalls and leaves the lock closed.
Through this failed first interaction, the policy infers that clockwise rotation is blocked, switches to turning counterclockwise, and successfully unlocks the mechanism.
Such traps emerge from physical properties exposed through interaction, including hidden mass distribution, surface friction, or transparent obstacles, as well as from cross-stage interference, where later actions undo completed subgoals~\cite{lu2026logiv}.
Furthermore, deploying robots in real homes demands autonomous task completion, as every scene reset in simulation corresponds to manual human intervention to rearrange the environment.
We prioritize in-attempt recovery and allow up to three attempts, resetting only when the execution budget runs out or the agent deems recovery impossible.

Existing benchmarks leave three gaps in evaluating how agents respond to their own failures.
First, episode-level success rates conflate clean execution with recovery within the same episode~\cite{diao2026liberoagent}.
Second, recovery benchmarks initialize policies from recorded failure states~\cite{liu2026liberorecover,li2026roborecover}, bypassing the stage where an agent makes an error, notices and diagnoses the failure, and decides how to recover.
Third, success over multiple attempts can improve through repeated sampling alone, obscuring whether higher performance comes from learning from failure or simply from additional retries.

To measure how robot policies learn from failure, we introduce \bench (Learning From Failure Benchmark), comprising 40 manipulation tasks built around verified physical and geometric traps (\cref{fig:teaser}).
Each task embeds a hidden cause that defeats standard, common-sense policies, requiring agents to detect failures and infer how to recover from their own observations.
The evaluator withholds success or failure feedback until the agent calls \texttt{finish} or exhausts its execution budget.
Specifically, 34 tasks each target a single failure cause, spanning execution offsets (L1), disturbed states requiring restoration (L2), infeasible strategies (L3), misjudged geometry (L4), and misjudged physics (L5), while six additional long-horizon tasks (L6) chain multiple traps across successive task stages.

Across all tasks, three Model Context Protocol (MCP) tools standardize observations, control primitives, and scoring, above which agent and robot harnesses are interchangeable.
Our fine-grained evaluator tracks trap activations and real-time subgoal progress to classify each run as clean success, in-attempt recovery, cross-attempt recovery, or failure.
We prioritize first-attempt success ($\Aone$), report three-attempt success ($\Athree$) as a secondary metric, and compare history-retained retries against memory-free resets ($\Delta_{\text{hist}}$).

Evaluating six multimodal agents across all 40 tasks (three instances each), GPT-6 Astra and Claude Opus~5.5 achieve the highest $\Aone$ at 50.0\% and 47.5\%, with remaining failures concentrated in L4--L6.
Progressive feedback identifies failure diagnosis as the main bottleneck for GPT-6 Astra and Opus~5.5---revealing the hidden cause largely resolves their errors---whereas weaker models struggle with both diagnosis and physical execution.
Across models, when later actions undo completed steps, agents seldom restore all lost progress within the attempt; on visually subtle failures, agents often notice anomalies yet rationalize them and claim success; and retaining history consistently outperforms memory-free retries.
Our prompt-only robot harness, LEAF-Lite, achieves the highest observed first-attempt and three-attempt success rates in our controlled harness comparison.

We make three main contributions.
\begin{itemize}[leftmargin=*,itemsep=1pt,topsep=2pt]
  \item \textbf{A benchmark and replay dataset for learning from failure.} We introduce \bench with 40 tasks spanning five failure causes and long-horizon recovery, and open-source over 6 million labeled replay frames with observations, robot actions, failure-recovery trajectories, and agent reasoning. Fine-tuning representative VLA and WAM policies on successful replays validates the training pipeline and reveals the limits of success-only imitation.
  \item \textbf{An evaluation protocol prioritizing autonomous recovery.} Our fine-grained evaluator separates clean success, in-attempt recovery, cross-attempt recovery, and failure. We prioritize first-attempt success ($\Aone$), report success within three attempts ($\Athree$) as a secondary metric, and pair history-retained retries with memory-free resets to isolate experience-driven gains, with interchangeable agent and robot harnesses above one MCP interface.
  \item \textbf{Empirical analysis of diagnosis and execution bottlenecks.} Across six agent-as-policy models, we identify diagnosis bottlenecks in frontier agents, execution barriers in weaker agents, and subgoal regressions and false completion claims. We quantify gains from retained history and validate LEAF-Lite, a prompt-only harness that applies these findings to guide recovery from failure.
\end{itemize}

%% file: sections/2_related.tex
\section{Related Work}
\label{sec:related}

\paragraph{Agent-as-policy benchmarks.}
Code-writing agents compose robot skills~\cite{liang2023codeaspolicies,huang2023voxposer}, while LIBERO-Agent and RobotWorld evaluate direct robot control across tasks and embodiments~\cite{diao2026liberoagent,yang2026robotworld}.
RoboQuest tests information gathering through interaction~\cite{liu2026roboquest}, but discloses that information is hidden and evaluates each episode once.
RoboARCade studies in-context adaptation and false success claims in long-horizon manipulation~\cite{anonymous2026roboarcade}.
\bench elicits agents' own failures, keeps causes hidden, and measures detection and recovery within and across attempts.

\paragraph{Failure recovery and reasoning.}
REFLECT and AHA study failure explanation from recorded rollouts~\cite{liu2023reflect,duan2025aha}, while LIBERO-RECOVER and RoboRecover evaluate recovery from stored failure states~\cite{liu2026liberorecover,li2026roborecover}.
The Physical Reasoning release of AGIBOT WORLD 2026 pairs high-level Q\&A with robot action trajectories~\cite{agibot2026agibotworld}; our replays include agents' recorded reasoning during failure and recovery.
LLF-Bench studies learning from language feedback~\cite{cheng2023llfbench}, and Reflexion retains verbal reflections across attempts~\cite{shinn2023reflexion}.
CoVer-VLA ranks candidate action chunks for alignment with the task instruction before execution~\cite{kwok2026scaling}.
RoboRetry conditions VLAs on prior failure contexts and separates structural retry cues from gains due to corrective language~\cite{liu2026roboretry}.
VRL-Bench compares retained verbal memory with memory-free retries in computer control~\cite{bai2026vrlbench}.
RoboMME evaluates how robot policies use memory in history-dependent manipulation tasks~\cite{dai2026robomme}.
\bench traces how agents cause, detect, and repair failures in visual manipulation, with event-level tracking and paired memory-free controls.

\paragraph{Agent and robot harnesses.}
Agent harnesses execute models and dispatch tools, while robot harnesses add planning, memory, or skills~\cite{lu2026logiv}.
RPent, SenseNova-RoboRSI, and EmbodiedEvo instantiate different forms of orchestration and experience reuse~\cite{zhang2026harnessvla,sensenova2026embodied,li2026embodiedevo}.
Their published systems differ in observations and action interfaces.
We compare controlled ports with shared tasks, tools, and scoring, varying one harness layer at a time with a fixed model (\cref{sec:harness}).

%% file: sections/3_benchmark.tex
\begin{figure*}[t]
  \centering
  \IfFileExists{figs/Protocol_reference.pdf}{\includegraphics[width=\linewidth]{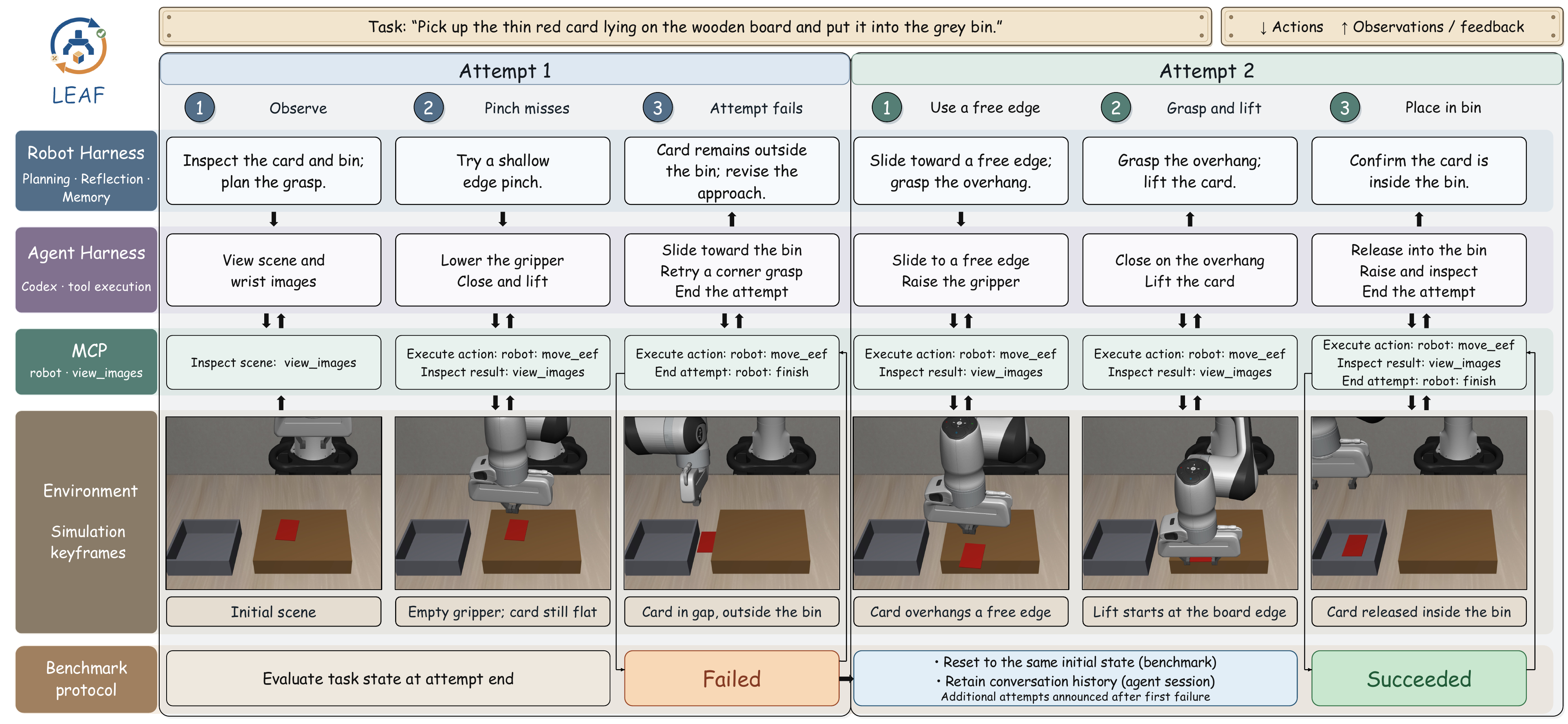}}{\fbox{\parbox{0.95\linewidth}{\centering\vspace{3em}protocol figure\vspace{3em}}}}
  \caption{\textbf{Protocol in action.} An L3 simulation rollout with GPT-6 Astra and Codex. The first attempt fails to place a thin card in the bin. After the same scene is reset with history retained, the agent slides the card to a free edge, grasps the overhang and places it in the bin. The agent harness dispatches MCP tools, and the privileged evaluator returns the attempt verdict.}
  \label{fig:gallery}
\end{figure*}

\section{The \bench Benchmark}
\label{sec:bench}

Measuring learning from failure requires three components: controlled tasks that induce persistent, self-generated failures (\cref{sec:taskdesign}), a closed-loop protocol supporting autonomous detection and repair (\cref{sec:protocol}), and event-level metrics decomposing recovery outcomes (\cref{sec:metrics}).

\subsection{Task design}
\label{sec:taskdesign}
\label{sec:principles}
\label{sec:tasks}

Every task in \bench contains a controlled \emph{trap}: a hidden cause that is hard to identify from static appearance alone and emerges only through physical interaction.
In \cref{fig:gallery}, a thin card lies flat on a wooden board, causing standard top-down grasps to fail.
To succeed, the agent must infer that the card is too thin to pinch from above, slide it to the edge of the board, and grasp the exposed overhang.
\bench comprises 40 manipulation tasks across six failure categories (\cref{fig:teaser}b).
The 34 single-cause tasks span five failure mechanisms: (L1) \emph{execution offsets} under camera-gripper misalignment; (L2) \emph{disturbed states} requiring object restoration; (L3) \emph{infeasible strategies} such as obstructed grasps; (L4) \emph{misjudged geometry} involving deceptive shapes or transparent obstacles; and (L5) \emph{misjudged physics} arising from hidden mass distribution, surface friction, or object compliance.
Six long-horizon tasks (L6) chain multiple traps across the physically coupled stages of a single manipulation task.

Each task pairs a LIBERO~\cite{liu2023libero} scene with a parametric instance sampler and a privileged evaluator, simulated with a Franka Panda arm and a parallel-jaw gripper in robosuite~\cite{zhu2020robosuite} and MuJoCo~\cite{todorov2012mujoco}.
Sampling three distinct instances per task yields 120 evaluation instances in total.
We validate every instance using two scripted reference policies.
First, a privileged policy with prior knowledge of the hidden cause completes the task, guaranteeing that every instance is physically solvable.
Second, a standard policy without access to this hidden cause triggers the trap and fails. To solve the task, the agent must infer the hidden cause from its own physical interactions with the scene.

\subsection{Protocol}
\label{sec:protocol}
\label{sec:interface}

\paragraph{Interface.}Agents interact through three MCP tools: \texttt{robot} for physical control, \texttt{shell} for executing custom analysis or control scripts, and \texttt{view\_images} for visual inspection.
The \texttt{robot} tool exposes three primitives: \texttt{move\_eef} commands an absolute end-effector target with an optional gripper action, \texttt{observe} captures updated camera views, and \texttt{finish} ends the attempt while requiring the agent to state whether the task is complete and explain its reason.
After every \texttt{robot} command, the server writes two $512 \times 512$ RGB frames (from a fixed third-person \texttt{agentview} camera and a wrist-mounted camera) and a proprioceptive state file into the workspace, closely matching standard real-world robot setups.
\paragraph{Attempts and History.}
An execution failure does not automatically terminate an attempt or trigger a scene reset.
Within an attempt, the scene changes only through the agent's own actions, so errors persist until the agent detects and repairs them; no evaluator intervention is provided.
An attempt ends when the agent calls \texttt{finish} or exhausts its execution budget, at which point the evaluator returns a success or failure verdict.
We measure in-attempt recovery and $\Aone$ within this protocol (\cref{sec:metrics}).

After an unsuccessful attempt, the environment resets to the same instance with the same random seed, preserving the layout and hidden cause, so a later attempt tests whether the agent has learned from its failure; an episode allows up to three attempts.
Six of the ten L2 tasks continue from the current state after an unsuccessful attempt. The agent must address disturbances such as knocked-off items, turned turntables, or interrupted sequences, all of which would be erased by resetting the scene to its initial state.

All attempts of an episode run in one agent session, so the \emph{history} is the agent's full context: its messages and reasoning as kept by the agent harness, all tool calls and robot replies, the images it viewed, each attempt's verdict, and its workspace files.
Retries are announced only after the first failure, so the first attempt matches that of a fresh agent; on the 34 tasks whose scene resets, a \emph{memory-free retry} starts a fresh agent in a new session after the scene has reset.

\paragraph{Unified Evaluation Architecture.}
Keeping tasks, MCP tools, and scoring fixed enables controlled comparisons across different system layers (\cref{fig:teaser}a).
At the top layer, an optional robot harness manages high-level planning, reflection, and memory.
Beneath it, an agent harness (such as Codex, Claude Code, or DeepSeek Harness) handles model execution and tool dispatch.
Standalone action policies, including VLA and WAM models, use a matched low-level action interface with identical observations, controllers, evaluators, and execution budgets, so we can compare these policies with agentic systems under the same protocol.

\paragraph{Isolation.}
To prevent coding agents from reward hacking or accessing privileged information such as task code, hidden trap parameters, and prior logs through filesystem inspection~\cite{openai2024o1systemcard,vonarx2025rewardhacking,zheng2026sweproverified}, each episode runs inside a network-isolated sandbox that exposes only a clean workspace directory, the proprioceptive state directory, and the MCP tool socket.
Preflight checks lock the harness prompt and tool schemas, and an automated audit verifies each tool call.

\subsection{Metrics}
\label{sec:metrics}

Throughout execution, the privileged evaluator logs trap activations alongside subgoal completions, regressions, and restorations.
Based on these event traces, each run is classified into one of four outcomes: (1) \emph{clean success} ($\text{CS}$), completing the task on the first attempt without any error or correction; (2) \emph{in-attempt recovery} ($\text{IR}$), repairing an error within the first attempt to reach completion; (3) \emph{cross-attempt recovery} ($\text{XR}$), failing on the first attempt and succeeding on the second or third attempt; and (4) \emph{failure} ($\text{F}$), failing across all allowed attempts.
Our primary metric is first-attempt success, $\Aone = \text{CS} + \text{IR}$, which reflects a policy's autonomous performance in real-world settings, whether it succeeds cleanly or recovers from errors.
As a secondary metric, we report three-attempt success, $\Athree = \Aone + \text{XR}$, to measure cumulative recovery across scene resets.
History gain ($\Delta_{\text{hist}}$) is the difference between $\Athree$ and three-attempt success with memory-free retries.

\subsection{Replay dataset and Policy Evaluation Suite}
\label{sec:data}

To support future research on failure-aware policy learning and evaluation across VLA and WAM paradigms, we release both our full evaluation replay dataset and a unified policy evaluation interface.
The replay dataset captures rich failure and recovery dynamics, including failed rollouts, in-attempt repairs, cross-attempt recoveries, and recorded agent reasoning traces.
It comprises 3303 trajectories (1043 successful and 2260 failed) across 2272 evaluation runs, totaling 6.45 million control steps and 89.6 simulated hours of 20\,Hz two-view images, proprioceptive states, and actions in LeRobot format.
Complementing the dataset, we open-source standardized evaluation interfaces and model adapters for standalone VLA and WAM policies, with representative models including $\pi_{0.5}$~\cite{physicalintelligence2025pi05}, DM0.5-MEM~\cite{dexmal2026dm05}, FastWAM~\cite{yuan2026fastwam}, and OpenWAM-$\alpha$~\cite{wang2026openwam} already integrated and benchmarked in our main results (\cref{sec:main}).
Together, these resources provide an open pipeline for training failure-aware embodied policies and systematically benchmarking how they detect and recover from their own manipulation failures.

%% file: sections/4_experiments_v2.tex
\section{Experiments}
\label{sec:experiments}

We first compare how different policy paradigms perform across all failure categories (\cref{sec:main}), and then follow an episode step by step to analyze why failures persist and how agents recover (\cref{sec:findings}). Finally, we compare external robot and agent harnesses under a shared protocol (\cref{sec:harness}). We then use the failure analysis to design LEAF-Lite, a prompt-only robot harness that guides physical probing, subgoal checks, and strategy revision across attempts (\cref{sec:real_world}).

\subsection{Setup}
\label{sec:setup}

We evaluate three policy paradigms using the shared scoring and attempt rules specified in \cref{sec:protocol}.
\begin{enumerate}[leftmargin=*,topsep=3pt,itemsep=2pt,parsep=0pt]
  \item \textbf{Agent-as-Policy.} We test six multimodal models in their standard agent harnesses with high reasoning effort and no external robot harness: GPT-6 Astra, GPT-6.1-Sol, DeepSeek V4.1 Flash and GPT-5.6-Sol in Codex, alongside Claude Opus~5.5 and Sonnet~5.5 in Claude Code. We evaluate 120 instances across 40 tasks with up to three attempts and retained history. After a failed first attempt on a task that resets, we also run a memory-free retry. 
  \item \textbf{Vision--Language--Action (VLA) and World Action Models (WAM).} To test closed-loop learning from LEAF replays, we fine-tune VLA ($\pi_{0.5}$, DM0.5-MEM) and WAM (FastWAM, OpenWAM-$\alpha$) models on successful trajectories from two instances per single-cause task. We test the held-out third instances through the matched low-level interface and omit L1, where the high-level MCP motion primitive injects the camera-gripper offset.
\end{enumerate}

\begin{table*}[!t]
\centering
\small
\setlength{\tabcolsep}{3pt}
\renewcommand{\arraystretch}{1.04}
\caption{\textbf{Main results across policy families.}
L1--L5 show first-attempt success ($\Aone$). Direct actions bypass the L1 command-level perturbation. Total tokens/run averages input tokens (including cached tokens) plus output tokens across all 120 instances, with up to three attempts per instance. Overall scores use 120 Agent-as-Policy and 31 held-out L2--L5 VLA/WAM instances.}
\label{tab:main}
\begin{tabular*}{\linewidth}{@{\extracolsep{\fill}}llrrrrrrrr@{}}
\toprule
& & \multicolumn{5}{c}{By level, $\Aone$ (\%) $\uparrow$}
  & \multicolumn{2}{c}{Overall (\%) $\uparrow$} & \\
\cmidrule(lr){3-7}\cmidrule(lr){8-9}
Policy & Agent harness & L1 & L2 & L3 & L4 & L5
& $\Aone$ & $\Athree$ & Total tokens/run $\downarrow$ \\
\midrule
\multicolumn{10}{@{}l}{\textit{Agent-as-Policy}} \\
\textbf{GPT-6 Astra} & Codex
& 88.9 & 70.0 & 91.7 & 20.0 & 33.3
& 50.0 & 86.7 & 4.3M \\
GPT-6.1-Sol & Codex
& 88.9 & 70.0 & 83.3 & 20.0 & 25.0
& 47.5 & 75.0 & 5.0M \\
Opus~5.5 & Claude Code
& 100.0 & 66.7 & 75.0 & 20.0 & 33.3
& 47.5 & 82.5 & 4.5M \\
Sonnet~5.5 & Claude Code
& 44.4 & 36.7 & 66.7 & 13.3 & 19.4
& 26.7 & 54.2 & 6.4M \\
DeepSeek V4.1 Flash & Codex
& 33.3 & 0.0 & 25.0 & 0.0 & 5.6
& 7.5 & 24.2 & 23.3M \\
GPT-5.6-Sol & Codex
& 0.0 & 0.0 & 25.0 & 0.0 & 16.7
& 7.5 & 20.0 & 8.9M \\
\midrule
\multicolumn{10}{@{}l}{\textit{Vision--Language--Action (VLA)}} \\
$\pi_{0.5}$ (SFT) & --
& -- & 0.0 & 25.0 & 0.0 & 8.3
& 6.5 & 22.6 & -- \\
\textbf{DM0.5-MEM} (SFT) & --
& -- & 10.0 & 50.0 & 20.0 & 8.3
& 16.1 & 19.4 & -- \\
\midrule
\multicolumn{10}{@{}l}{\textit{World Action Model (WAM)}} \\
FastWAM (SFT) & --
& -- & 10.0 & 25.0 & 0.0 & 0.0
& 6.5 & 6.5 & -- \\
\textbf{\boldmath OpenWAM-$\alpha$} (SFT) & --
& -- & 0.0 & 50.0 & 20.0 & 0.0
& 9.7 & 19.4 & -- \\
\bottomrule
\end{tabular*}
\end{table*}

\subsection{Main results}
\label{sec:main}

\paragraph{Frontier agents lead in the completed evaluations.}
GPT-6 Astra and Claude Opus~5.5 lead overall performance, reaching 50.0\% and 47.5\% on the first attempt ($\Aone$) and 86.7\% and 82.5\% within three attempts ($\Athree$), well ahead of Sonnet~5.5 (26.7\% $\Aone$) and DeepSeek V4.1 Flash (7.5\% $\Aone$, \cref{tab:main}).
Fine-tuned VLA and WAM policies reach up to 50.0\% $\Aone$ on obstructed grasps (L3) by learning feasible strategies from successful replays, verifying our closed-loop training and evaluation pipeline. Overall $\Aone$ reaches at most 16.1\%, with maxima of 20.0\% on L4 and 8.3\% on L5. \hyperref[e.2.-training-with-outcome-labeled-failure-rollouts]{Appendix~\ref*{e.2.-training-with-outcome-labeled-failure-rollouts}} examines closed-loop execution failures and the limits of recovery under the tested fine-tuning setup.

\begin{table}[!b]
\centering
\small
\setlength{\tabcolsep}{3pt}
\caption{\textbf{Agent outcomes.}
Percentages over 120 instances (L1--L6) for six agents without a robot harness.
CS is clean success, IR is in-attempt recovery, XR is cross-attempt recovery, and F is failure across all allowed attempts.}
\label{tab:outcomes}
\begin{tabular*}{\columnwidth}{@{\extracolsep{\fill}}lrrrr@{}}
\toprule
Agent & CS (\%) & IR (\%) & XR (\%) & F (\%) \\
\midrule
\textbf{GPT-6 Astra} & 6.7 & 43.3 & 36.7 & 13.3 \\
GPT-6.1-Sol & 5.0 & 42.5 & 27.5 & 25.0 \\
Opus~5.5 & 15.0 & 32.5 & 35.0 & 17.5 \\
Sonnet~5.5 & 3.3 & 23.3 & 27.5 & 45.8 \\
DeepSeek V4.1 Flash & 0.0 & 7.5 & 16.7 & 75.8 \\
GPT-5.6-Sol & 0.0 & 7.5 & 12.5 & 80.0 \\
\bottomrule
\end{tabular*}
\end{table}

\paragraph{First attempts break on hidden geometry and physics, and on chained traps.}
For the two leading agents, first-attempt success reaches 67\%--100\% on execution offsets (L1), disturbed states (L2), and infeasible strategies (L3), yet drops to 20\% on misjudged geometry (L4), 33\% on misjudged physics (L5), and 22\%--28\% on chained tasks (L6).
Within three attempts, GPT-6 Astra and Opus~5.5 recover to 53\%--86\% on L4--L6, whereas weaker agents stall at 11\% on L6. Execution deviations and blocked grasps are largely fixed within the first try, whereas hidden properties and multi-stage traps require adapting across attempts.

\paragraph{Most first-attempt successes involve in-attempt repair.}
Clean, error-free runs are rare even for top agents: 87\% of GPT-6 Astra's and 68\% of Opus~5.5's first-attempt successes contain an error repaired within the attempt (\cref{tab:outcomes}).
In \cref{tab:main}, GPT-6 Astra reaches 50.0\% $\Aone$ at 4.3M total tokens/run, and DeepSeek V4.1 Flash reaches 7.5\% at 23.3M.

\subsection{Findings}
\label{sec:findings}

We use controlled feedback and event traces to explain why some failures can be repaired and others persist. We first separate diagnosing a trap from executing a repair by varying what the agent knows about the failure (Findings~1--2). We then examine whether agents restore undone subgoals and whether their completion claims match the final scene (Findings~3--4). Finally, we compare retries with and without retained history after a shared first attempt to measure the benefit of earlier interactions (Finding~5).

\begin{figure}[!ht]
  \centering
  \captionsetup{skip=6pt}
  \includegraphics[width=\columnwidth,trim=0bp 0bp 250bp 0bp,clip]{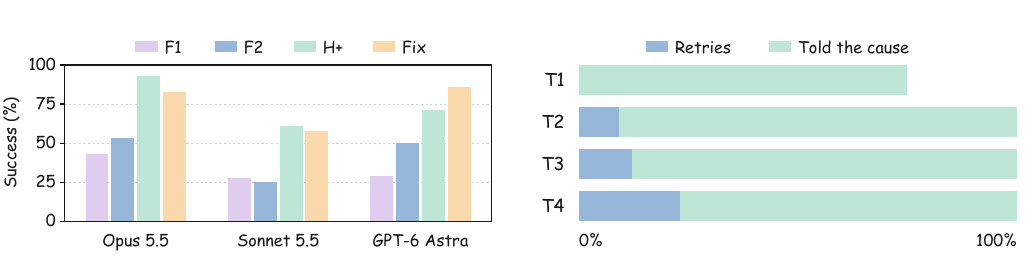}
  \caption{\textbf{Hidden-cause feedback helps agents recover.}
  Success of fresh agents restarting from the same failed first attempts with different feedback: failure alone (F1), the measured error (F2), the hidden cause (H+), or the fix (Fix).}
  \label{fig:diagnosis}
\end{figure}

\finding{1}{Diagnosing the hidden cause is the main bottleneck for frontier models.}
Every single-cause instance left unsolved by Opus~5.5 falls in L4 or L5. Starting from failed first attempts on these traps, we restart fresh agents while progressively adding feedback (\cref{fig:diagnosis}).
Told only that the attempt failed (F1), agents repeat the same flawed choice, such as never removing the clear lid on Clear Cover (\hyperref[tab:s2]{Table~\ref*{tab:s2}}). When GPT and Opus miss the cause, in 51\% of such attempts they blame motion slips and tweak speed, force, or distance, and in 33\% they treat the anomaly as part of the scene.
Adding the measured error (F2) improves recovery for GPT-6 Astra and Opus~5.5. Revealing the hidden cause (H+) yields larger gains, raising recovery from 43\% to 93\% for Opus~5.5 and from 29\% to 71\% for GPT-6 Astra, matching most of the gain from giving the explicit repair (Fix). Frontier models can already execute the repair once the cause is known, making causal diagnosis the primary bottleneck for autonomous recovery.

\finding{2}{Weaker models struggle with both diagnosis and physical execution.}
The causal-feedback test (H+) also separates models by execution ability: when told the hidden cause, Opus~5.5 succeeds in 93\% of runs whereas Sonnet~5.5 reaches 61\%. Execution logs trace this gap back to basic manipulation, where Sonnet holds the object on its first grasp less than half as often as Opus (30\% vs.\ 68\%).
Low initial precision can be offset by active repair---GPT-6 Astra holds only 40\% of first grasps yet matches Opus in $\Aone$ by fixing its own slips twice as often within the attempt. For weaker models like Sonnet, missed causes and execution errors account for nearly equal shares of multi-attempt failures, showing that causal hints alone fall short when motor control introduces new errors during repair.

\finding{3}{Agents rarely restore completed steps they accidentally undo.}
Unlike diagnosing a fixed hidden cause (Findings 1–2), recovering from subgoal regressions requires continuously tracking and restoring previously completed progress. Across the 13 tasks that log subgoals, later actions break an earlier completed step---such as knocking placed blocks off their marks while closing a lid---in roughly a third of attempts for Opus~5.5 (36\%) and half for GPT-6 Astra and Sonnet~5.5. Once a step is undone, agents restore all lost progress in at most a third of those attempts (10\%--32\%).
Trajectory logs show that once agents check off a subgoal in their plan, they keep working on the remaining objects and seldom look back after later contact disturbances, leaving cross-stage interference unrepaired.

\begin{figure}[!ht]
  \centering
  \includegraphics[width=\columnwidth]{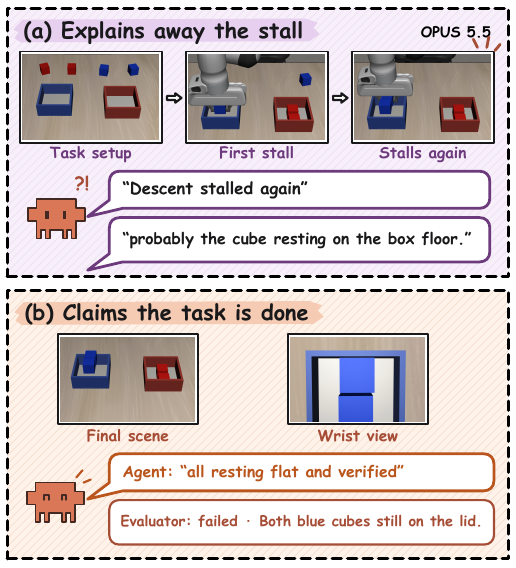}
  \caption{\textbf{From a noticed anomaly to a false completion claim.}
  On Clear Cover, Opus~5.5 attributes a stalled descent to the box floor and releases the cube. It later claims that placement is verified, but both blue cubes remain on the transparent lid.}
  \label{fig:clear-cover-claim}
\end{figure}

\phantomsection
\finding{4}{Agents notice subtle anomalies yet explain them away and claim success.}
\label{sec:remains}
On visually subtle failures in Clear Cover and Fit Container, agents misjudge task completion and claim success on 85\% of failed attempts. Reasoning logs show that in 69\% of these false claims by GPT-6 Astra and Opus~5.5, the agent notices an execution anomaly, explains it away as normal, and reports completion.
For example, on Clear Cover (\cref{fig:clear-cover-claim}), Opus~5.5 detects repeated stalled descents, assumes the cube has reached the box floor, and releases it---later claiming the cubes are \textquotedblleft all resting flat and verified\textquotedblright{} while both remain on the transparent lid.
Across these false claims, just 14\% inspect the scene after the final release. Even prompting Opus~5.5 to verify every goal condition lifts success from 0\% to 13\% (versus 88\% when told the cause), showing that resolving visually ambiguous traps requires forming a hypothesis and actively probing uncertain surfaces with the gripper.

\begin{figure}[!ht]
  \centering
  \captionsetup{skip=5pt}
  \includegraphics[width=\linewidth,trim=0 4bp 0 0,clip]{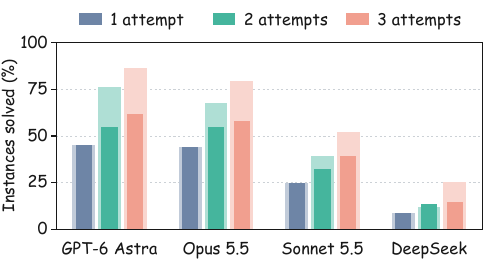}
  \caption{\textbf{History against memory-free retries.}
Each group shows cumulative success after one, two, and three attempts on instances with scene resets. Wide, light bars show runs with retained history. Narrow, dark bars show runs with a fresh agent for each retry. Both conditions share the first attempt, so the effect of retaining history is measured by comparing the two retry conditions.}
  \label{fig:history}
\end{figure}

\finding{5}{Retaining history consistently improves recovery over memory-free retries.}
Keeping interaction history across attempts substantially boosts recovery beyond repeated sampling alone (\cref{fig:history}). While memory-free retries raise success by avoiding earlier slips---lifting GPT-6 Astra from 45\% to 62\% and Opus~5.5 from 44\% to 58\%---retaining history pushes the same models to 86\% and 79\%, with history gains of 24.5 points for GPT-6 Astra, 21.6 for Opus~5.5, 12.7 for Sonnet~5.5, and 10.8 for DeepSeek V4.1 Flash.
Across paired runs, 20\% of single-cause instances succeed only with history, compared to just 1\% without it, confirming that agents use prior failed interactions to adapt their strategies across attempts.

\subsection{External harnesses under one protocol}
\label{sec:harness}

We vary one harness layer at a time while holding tasks, tools, and scoring fixed (\cref{tab:harness,tab:agent-harness}). \Cref{sec:real_world} describes the design and evaluation of LEAF-Lite.

\begin{table*}[!t]
\centering
\small
\captionsetup{skip=5pt}
\caption{\textbf{Robot harness comparison.} All methods run gpt-6.1-sol in Codex at high reasoning effort. Success rates cover 40 tasks from L1--L6, with three instances per task (120 instances total). All rates are percentages. Robot harness ports follow our protocol.}
\label{tab:harness}
\setlength{\tabcolsep}{4pt}
\begin{tabular*}{\textwidth}{@{\extracolsep{\fill}}lrrrrr@{}}
\toprule
Robot harness & $\Aone$ & $\Athree$ & Total tokens/run & Subgoal Restoration Rate $\uparrow$ & False Completion Claim Rate $\downarrow$ \\
\midrule
None (direct) & 47.5 & 75.0 & 5.0M & 25 & 22 \\
RPent~\cite{zhang2026harnessvla} & 48.3 & 84.2 & 3.6M & 33 & 28 \\
SenseNova-RoboRSI~\cite{sensenova2026embodied} & 45.8 & 75.8 & 5.2M & 25 & 22 \\
EmbodiedEvo~\cite{li2026embodiedevo} & 49.2 & 79.2 & 5.3M & 50 & 28 \\
\textbf{LEAF-Lite} (ours) & \textbf{52.5} & \textbf{90.0} & \textbf{3.5M} & \textbf{60} & \textbf{14} \\
\bottomrule
\end{tabular*}
\end{table*}

\smallskip
\noindent\textbf{Certain structured robot harnesses boost cross-attempt recovery by targeting diagnosis bottlenecks.}
With gpt-6.1-sol in Codex, the ported robot harnesses mainly improve success across attempts. First-attempt success ($\Aone$) ranges from 45.8\% to 49.2\%, against 47.5\% for the direct baseline. Three-attempt success ($\Athree$) rises from 75.0\% for the baseline to 79.2\% with EmbodiedEvo~\cite{li2026embodiedevo} and 84.2\% with RPent~\cite{zhang2026harnessvla} (\cref{tab:harness}).
To understand what drives this improvement in $\Athree$, we examine RPent's orchestration design and find that its core requirements directly align with our findings in \cref{sec:findings}.
First, requiring every retry to modify an explicit, named physical lever breaks strategy repetition under binary feedback (Finding~1). Logging strictly factual outcomes even when the cause remains unknown reduces causal misattribution and anomaly rationalization (Findings~1 and~4).
Second, concise factual archives retain actionable history (Finding~5) across fresh sessions. RPent uses 3.6M total tokens per run, compared with 5.0M for the direct baseline.
EmbodiedEvo likewise structures experience reuse and reflection across attempts, addressing the bottlenecks that keep agents from revising their strategies after an unsuccessful attempt on the same task instance.

\begin{table}[!t]
\centering
\small
\captionsetup{skip=5pt}
\caption{\textbf{Agent harness comparison.} High reasoning effort and no robot harness. Each setting covers 120 instances across 40 tasks.}
\label{tab:agent-harness}
\setlength{\tabcolsep}{4pt}
\begin{tabular*}{\columnwidth}{@{\extracolsep{\fill}}lrrr@{}}
\toprule
Agent harness & $\Aone$\,\% & $\Athree$\,\% & Total tokens/run \\
\midrule
\multicolumn{4}{@{}l}{\textit{DeepSeek V4.1 Flash}} \\
Codex & 7.5 & 24.2 & 23.3M \\
Claude Code & 7.5 & 25.0 & 19.4M \\
DeepSeek Harness & 7.5 & 24.2 & 32.5M \\
\midrule
\multicolumn{4}{@{}l}{\textit{Sonnet~5.5}} \\
Codex & 25.0 & 51.7 & 7.5M \\
Claude Code & 26.7 & 54.2 & 6.4M \\
DeepSeek Harness & 24.2 & 49.2 & 7.3M \\
\bottomrule
\end{tabular*}
\end{table}

\par\smallskip
\noindent\textbf{Success varies little across agent harnesses.}
With the model fixed and no robot harness, DeepSeek V4.1 Flash reaches 7.5\% $\Aone$ across all three harnesses and 24.2--25.0\% $\Athree$ (\cref{tab:agent-harness}). Sonnet~5.5 reaches 24.2--26.7\% $\Aone$ and 49.2--54.2\% $\Athree$. Claude Code achieves the highest observed $\Athree$ for both models and uses the fewest tokens per run. Across harnesses, token use ranges from 19.4M to 32.5M for DeepSeek and from 6.4M to 7.5M for Sonnet.

\subsection{Closing the Loop with LEAF-Lite}
\label{sec:real_world}

Finally, we test whether prompt-level guidance alone can address the recovery bottlenecks identified in \cref{sec:findings}. Keeping the agent harness and MCP tools fixed, we distill these findings into \textbf{LEAF-Lite}, a minimalist robot harness built on three principles. (1) executing small, reversible probes to test physical assumptions before committing to irreversible actions (Findings~1 and~4); (2) tracking completed subgoals by re-checking every previously completed subgoal upon finishing a new one, and verifying all goal states post-release before calling \texttt{finish} (Findings~3 and~4); and (3) recording a brief factual note in the workspace after a failed attempt and changing at least one named, decisive physical choice---rather than tweaking speed, force, or small offsets---on the next retry (Findings~1 and~5). Across all 120 instances, LEAF-Lite reaches 52.5\% $\Aone$ and 90.0\% $\Athree$ at 3.5M total tokens per run with gpt-6.1-sol in Codex (\cref{tab:harness}). It achieves the highest observed success rates in this comparison and uses about 30\% fewer tokens per run than the direct baseline. \hyperref[d.-statistical-reliability]{Appendix~\ref*{d.-statistical-reliability}} reports the paired \(A_3\) comparison with RPent and its statistical scope.

%% file: sections/5_conclusion.tex
\newpage
\section{Conclusion}
\label{sec:conclusion}

We introduced LEAF, a benchmark of 40 trap-driven manipulation tasks paired with a unified MCP interface, a fine-grained event evaluator, and memory-free controls to measure how robot policies detect, diagnose, and repair their own failures. Across the evaluated agent-as-policy models, retaining interaction history boosts three-attempt success by 11--25 percentage points over memory-free retries. The VLA and WAM policies in our evaluation show limited recovery after fine-tuning exclusively on successful manipulation trajectories. Even frontier multimodal agents remain bottlenecked by causal diagnosis, unrecovered subgoal regressions, and false completion claims on subtle anomalies, while weaker models face coupled execution barriers. \newpage
LEAF-Lite turns these findings into prompt-level rules for testing physical hypotheses, rechecking completed subgoals, and revising actions between attempts. In the controlled robot-harness comparison, it achieves the highest observed first-attempt and three-attempt success rates, uses fewer tokens per run than the direct baseline, and requires no extra orchestration code.

\vspace{8.25bp}
\textbf{Limitations and future work.} First, LEAF currently focuses on tabletop manipulation with a single arm and a parallel-jaw gripper, leaving out failure modes specific to bimanual coordination, dexterous hands, and mobile manipulation---such as inter-arm collisions, multi-finger in-hand slips, and coupled navigation-manipulation errors. Second, although agent-as-policy models adapt effectively to hidden traps, multi-turn tool-based reasoning incurs millions of tokens per episode and substantially higher inference latency than reactive controllers, limiting deployment on dynamic or time-critical tasks. Future work will extend LEAF to bimanual, dexterous, and mobile embodiments. We will use our 6.45M-step replay dataset to study whether VLA and WAM policies can learn active physical probing and closed-loop self-correction from failed rollouts, in-attempt repairs, and agent reasoning traces while preserving the low latency of direct action prediction.

%% file: sections/6_llm_usage.tex
\section*{LLM Usage Statement}
OpenAI Codex was used as a writing assistant during manuscript preparation, primarily for English-language polishing, sentence and paragraph restructuring, and improving clarity and concision across the manuscript and supplementary material.
It also assisted with source lookup and LaTeX formatting and layout.
These revisions were guided by author-provided content and instructions, with the authors selecting and refining the final wording.

Claude Opus~5.5 assisted in annotating failure causes and false success claims; all annotations were reviewed by humans.
The models evaluated as robot policies and their experimental configurations are described in \cref{sec:bench,sec:experiments}.
The authors led the research and retain full responsibility for the scientific content, reported results, factual accuracy, citations, and final manuscript.

%% file: supplementary.tex
\maketitlesupplementary
\appendix
\setcounter{figure}{0}
\setcounter{table}{0}
\renewcommand{\thefigure}{S\arabic{figure}}
\renewcommand{\thetable}{S\arabic{table}}
\renewcommand{\theHfigure}{supp.\arabic{figure}}
\renewcommand{\theHtable}{supp.\arabic{table}}
\setcounter{tocdepth}{1}
\raggedbottom
\setlength{\LTpre}{5pt}
\setlength{\LTpost}{8pt}

\begin{strip}
\noindent\begin{minipage}[t]{\columnwidth}

\section*{Appendix Contents}
\begingroup
\setlength{\parskip}{5pt}
\noindent\hyperref[a.-benchmark-task-specifications-and-metric-definitions]{\textbf{A. Benchmark Tasks}\dotfill\pageref*{a.-benchmark-task-specifications-and-metric-definitions}}\par
\noindent\hyperref[b.-detailed-analysis-of-failure-recovery-and-episode-anatomy]{\textbf{B. Failure and Recovery Analyses}\dotfill\pageref*{b.-detailed-analysis-of-failure-recovery-and-episode-anatomy}}\par
\noindent\hyperref[c.-from-failure-analysis-to-recovery-strategies]{\textbf{C. Recovery Strategies}\dotfill\pageref*{c.-from-failure-analysis-to-recovery-strategies}}\par
\noindent\hyperref[d.-statistical-reliability]{\textbf{D. Statistical Reliability}\dotfill\pageref*{d.-statistical-reliability}}\par
\noindent\hyperref[e.-learned-action-policies-implementation-and-ablations]{\textbf{E. Learned Action Policies}\dotfill\pageref*{e.-learned-action-policies-implementation-and-ablations}}\par
\noindent\hyperref[f.-replay-dataset-documentation]{\textbf{F. Replay Dataset}\dotfill\pageref*{f.-replay-dataset-documentation}}\par
\endgroup

\end{minipage}\hfill
\begin{minipage}[t]{\columnwidth}

\section{Benchmark Task Specifications}\label{a.-benchmark-task-specifications-and-metric-definitions}

LEAF comprises 40 manipulation tasks simulated in MuJoCo \cite{todorov2012mujoco} and robosuite \cite{zhu2020robosuite} on top of LIBERO scenes \cite{liu2023libero}: 34 single-cause tasks spanning five failure mechanisms (L1--L5) and six long-horizon tasks (L6) that chain three physically coupled traps across successive manipulation stages. \textbf{\hyperref[fig:s1]{Figure S1}} provides a visual overview across all 40 tasks, and \textbf{\hyperref[tab:s1]{Table S1}} lists the primary failure cause, qualitative physical trap, and verbatim language instruction provided to the agent for every task.

Task instructions specify only the manipulation goal and operational rules, and \textbf{\hyperref[tab:s1]{Table S1}} (Instruction column) shows the exact prompt fed to the agent without disclosing the hidden trap. While each task is built around one primary obstacle---or three chained obstacles in L6---execution slips during closed-loop control can introduce secondary disturbances during rollout that the agent must also resolve.

\end{minipage}
\par\vspace{12pt}

\centering
\includegraphics[width=1\linewidth,height=\textheight,keepaspectratio]{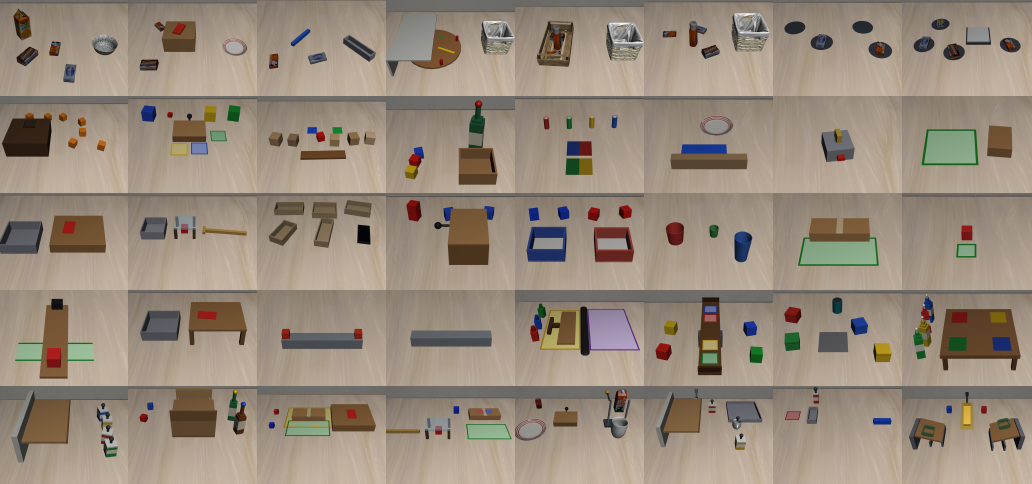}
\captionsetup{hypcap=false}\captionof{figure}{\textbf{Overview of all 40 tasks in LEAF.} Representative initial scenes across the 34 single-cause tasks ({L1}--{L5}) and six multi-trap long-horizon tasks ({L6}), arranged in a \(5 \times 8\) grid. Tasks follow the row-wise order of Table S1.}\label{fig:s1}
\end{strip}
\onecolumn

{\fontsize{9}{10}\selectfont
\setlength{\tabcolsep}{4pt}
\renewcommand{\arraystretch}{1.12}
\begin{longtable}{@{}
  >{\raggedright\arraybackslash}p{(\linewidth - 6\tabcolsep) * \real{0.1400}}
  >{\raggedright\arraybackslash}p{(\linewidth - 6\tabcolsep) * \real{0.0700}}
  >{\raggedright\arraybackslash}p{(\linewidth - 6\tabcolsep) * \real{0.2600}}
  >{\raggedright\arraybackslash}p{(\linewidth - 6\tabcolsep) * \real{0.5300}}@{}}
\caption{\textbf{Specifications of all 40 LEAF tasks.} Each task is listed with its primary failure category, qualitative physical trap, and verbatim instruction given to the agent. The hidden trap is never disclosed in the prompt. \textsuperscript{$\dagger$} denotes one of the six {L2} tasks that continue from the disturbed scene state after an unsuccessful attempt rather than resetting to the initial state. \textsuperscript{$\ddagger$} denotes one of the 13 tasks included in the subgoal-restoration analysis in Finding 3 of the main paper.}\label{tab:s1}\\
\toprule\noalign{}
\begin{minipage}[b]{\linewidth}\raggedright
Task
\end{minipage} & \begin{minipage}[b]{\linewidth}\raggedright
Cause
\end{minipage} & \begin{minipage}[b]{\linewidth}\raggedright
Trap
\end{minipage} & \begin{minipage}[b]{\linewidth}\raggedright
Instruction
\end{minipage} \\
\midrule\noalign{}
\endfirsthead
\toprule\noalign{}
\begin{minipage}[b]{\linewidth}\raggedright
Task
\end{minipage} & \begin{minipage}[b]{\linewidth}\raggedright
Cause
\end{minipage} & \begin{minipage}[b]{\linewidth}\raggedright
Trap
\end{minipage} & \begin{minipage}[b]{\linewidth}\raggedright
Instruction
\end{minipage} \\
\midrule\noalign{}
\endhead
\bottomrule\noalign{}
\endlastfoot
Biased Place & L1 & A fixed horizontal offset is added to every commanded position, so a grasp at the seen position misses. & Pick up the cream cheese and put it into the bowl. \\
Height Bias & L1 & A fixed upward offset is added to every commanded height, so the gripper stops short of the tile. & Pick up the red tile from the top of the wooden block and put it on the plate. \\
Yaw Bias & L1 & A fixed offset is added to every commanded wrist yaw and does not appear in the reported yaw. & Pick up the blue bar and put it into the holder. \\
Turntable Restore\textsuperscript{$\dagger$} & L2 & Reaching the butter requires turning the turntable; its start angle is shown only in the initial images. & Put the butter into the basket. The butter is on the turntable, under the shelf. You can turn the turntable by its red pegs. When you are done, the turntable must be turned exactly the way it is now. \\
Knock Restore\textsuperscript{$\dagger$} & L2 & Lifting the ketchup knocks a neighboring box off the tray. & Move the ketchup from the tray into the basket. Everything else must stay on the tray. \\
Restore Layout\textsuperscript{$\dagger$} & L2 & Lifting the ketchup displaces one of the boxes; its original pose is shown only in the initial images. & Put the ketchup into the basket. Every other item on the table must stay exactly where it is now, in the same position and orientation. \\
Swap Interrupt\textsuperscript{$\dagger$} & L2 & During a carry the item slips from the gripper and lands near a different pad. & Swap the positions of the cream cheese and the butter: each must end up on the round pad where the other one started. You may use the empty pads as temporary spots. \\
Sequence Resume\textsuperscript{$\dagger$} & L2 & Partway through, a stop returns the item to its pad and the arm to home, leaving the scene looking almost unchanged. & Weigh the four grocery items one at a time, in this order: ⟨order⟩. To weigh an item, put it on the grey scale and let go of it; the scale beeps when it has recorded the reading. Then put the item back on its own round pad before you pick up the next item. Weigh every item exactly once: the scale logs every reading, and the log cannot be edited. \\
Count into Box\textsuperscript{$\dagger$} & L2 & Partway through, one cube slips from the gripper back onto the table; the box contents cannot be seen. & Put exactly ⟨N⟩ of the orange cubes into the dark box, through the opening in its lid. Cubes cannot be taken back out of the box. \\
Lid Box\textsuperscript{$\ddagger$} & L2 & The opened lid falls back far enough to sweep blocks placed behind the box off their marks. & Put each of the three coloured blocks on the square mark of the same colour on the table. Then open the lid of the wooden box (lift it by the black knob on its front edge), put the red cube inside the box, and close the lid. \\
Bridge\textsuperscript{$\ddagger$} & L2 & Fetching the red cube after the plank is placed brings the hand into the bridge. & Build a small bridge. First stack two of the wooden blocks on the blue square, then two on the green square, to make two pillars; then lay the long plank across the tops of both pillars; finally put the red cube on the middle of the plank. The finished bridge must stand on its own, with the gripper open and not touching anything. \\
Bystander Bottle\textsuperscript{$\ddagger$} & L2 & A low transfer brings the arm into the tall, light bottle. & Put the red, yellow and blue cubes into the open wooden box. Do not move anything else: the tall glass bottle must be left standing upright exactly where it is now. (The bottle has a round red cap; held by its cap, the bottle hangs straight down.) \\
Nudge Row\textsuperscript{$\ddagger$} & L2 & Opening the fingers sideways on release pushes a neighboring cup off its mat. & Stand each of the four small cups on the mat of the same colour. The four coloured mats lie on the table in two rows of two. At the end every cup must stand upright with its whole base on its own mat (not touching another mat or the table around it), and the gripper must be open and clear of the cups. \\
Wall-Flush Box & L3 & The box rests against a wall with no room for a finger, so a direct grasp is blocked. & Put the long box on the plate. \\
Key Turn & L3 & The key turns in only one direction; the other direction hits a hard stop. & Unlock the lock on the grey box: grasp the flat head of the brass key from above and turn the key a quarter turn (90 degrees). The red bolt on the box slides in when it is unlocked. \\
Too-Wide Push & L3 & The box is slightly wider than the maximum gripper opening. & Put the brown box completely inside the green square marked on the table. \\
Card Edge & L3 & The card is too thin to pinch while it lies flat on the board. & Pick up the thin red card lying on the wooden board and put it into the grey bin. \\
Under-Table Tool & L4 & A transparent panel, not rendered in either camera, closes one side of the low table. & Put the red cube into the gray bin. \\
Fit Container & L4 & Clear end caps make the phone longer than it looks, so the box that looks large enough is too short. & Put the phone flat inside a box that it fits into completely. \\
False-Bottom Lid\textsuperscript{$\ddagger$} & L4 & The box has a false bottom that looks identical to a deep one, so the upright block keeps the lid from closing. & Put the red block and the two blue cubes into the wooden box and close its lid. The lid opens by lifting the black knob on its edge. \\
Clear Cover\textsuperscript{$\ddagger$} & L4 & A nearly invisible clear lid covers one box, so cubes released above it rest on the lid. & Put the two red cubes into the red box and the two blue cubes into the blue box. \\
Nest Order & L4 & The tallest cup is not the widest, so nesting by height jams. & Nest the red, green and blue cups one inside another. \\
Off-Center Push & L5 & The box's center of mass is off-center, so a centered push rotates it. & Slowly push the long brown box with the closed gripper so that it ends up completely inside the green rectangle ahead of it (farther from the robot). \\
Slide to Target & L5 & The cube--table friction is unknown, so the push speed that stops the cube in the square must be found. & Push the red cube straight away from the robot so that it slides and comes to rest in the green square marked on the table. On the side of the square nearest the robot and on its left and right, no part of the cube may lie on or beyond the green border line. On the far side, away from the robot, the cube may lie on the border line, but no part of it may stick out beyond that line. The cube must slide the last part of the way on its own: after it leaves the gripper it has to slide at least 3 cm, with nothing touching it, before it comes to rest in the square. \\
Tablecloth Pull & L5 & The cube's friction is unknown, so a pull at ordinary speed leaves the cube outside the band. & Pull the wooden board out from under the red cube so that the cube ends up standing on the table completely inside the green band. Touch only the board and its black handle, never the cube. Only one pull is allowed per attempt: grasp the black handle, pull the board out with ONE move command, then release it. Any other move that drags, pushes or nudges the board, before or after the pull, counts as another pull. \\
Card on Shelf & L5 & One end of the card is weighted; slid out heavy end first, the card tips off the shelf. & Pick up the thin red card lying on the wooden shelf and put it into the grey bin. \\
Loaded Beam\textsuperscript{$\ddagger$} & L5 & An internal weight shifts the bar's center of mass, so a lift at the middle tilts it and the blocks slide off. & Lift the metal bar together with the two small red blocks resting near its ends, and hold it level (tilted less than 8 degrees) with its lowest point at least 10 cm above the table. Both blocks must stay on the bar near its ends. Keep holding the bar when you finish. \\
Slippery Lift & L5 & An internal weight shifts the bar's center of mass, so a grasp at the middle lets the bar tilt. & Pick up the metal bar and hold it level (tilted less than 8 degrees), with its lowest point at least 10 cm above the table. Keep holding it when you finish. \\
Tray Delivery\textsuperscript{$\ddagger$} & L5 & Some identical-looking bottles are top-heavy and topple when the loaded tray is pushed over the strip. & Load the red, green and blue bottles onto the wooden tray: stand each bottle upright on the tray while the tray is still inside the yellow loading area. Only after all three bottles are on the tray, move the tray with the bottles on it over the black threshold strip into the purple delivery area. At the end the whole tray, including its handle, must be inside the purple area with all three bottles standing upright on it. \\
Balance Board\textsuperscript{$\ddagger$} & L5 & The cubes have unequal hidden masses, so placing them in the wrong order tips the plank. & Put each coloured cube on the square mark of the same colour on the long plank. The plank lies loose across a narrow block under its middle. At the end all four cubes must be on their marks and the plank must still rest level on the block. \\
Tidy Stack\textsuperscript{$\ddagger$} & L5 & One identical-looking box is an empty carton that collapses under load. & Tidy up the table: stack the four storage boxes (red, green, blue, yellow) into a single pile on the grey mat, in any order you like, and then put the cup upright on top of the pile. At the end each box must rest flat on the box below it (the bottom one on the mat), the cup must stand on the top box, and the gripper must be open and clear of everything. \\
Wobbly Table\textsuperscript{$\ddagger$} & L5 & One table leg is short and one bottle is heavy; placing it rocks the table and knocks over standing bottles. & Put each of the four bottles on the spot of its own colour on the small wooden table. At the end every bottle must stand upright on its spot, the gripper must be open and clear of the bottles, and the small table must still stand where it is now. (Each bottle has a square neck with a small round white stopper on top. Gripped on the neck, a bottle stays rigid in the fingers; gripped by the stopper alone, it swings freely and hangs straight down.) \\
Shelf Droop\textsuperscript{$\ddagger$} & L5 & The shelf bracket is loose and gives way under the load of the jars. & Put the four jars (red, green, blue, yellow) on the shelf. At the end all four must be standing upright on the shelf board, and the gripper must be open and clear of the jars and the shelf. (Each jar has a round knob on its lid; held by the knob, a jar hangs straight down.) \\
Drawer Cabinet\textsuperscript{$\ddagger$} & L5 & The drawer closes itself on a spring, jerking the cabinet and toppling bottles standing on it. & Tidy up the table: stand the green bottle and the brown bottle upright on top of the small wooden cabinet, put the red cube and the blue cube into the cabinet's open drawer (it is pulled out towards the robot), and close the drawer. When you finish, both bottles must stand upright on the cabinet top, both cubes must be in the closed drawer, the cabinet must still stand where it is, and the gripper must be open and not touching anything. (Each bottle has a round cap; held by its cap, a bottle hangs straight down.) \\
Pack and Deliver & L6 (L1, L3, L5) & A command offset makes grasps miss; the thin card cannot be pinched flat; the loaded tray's off-center mass rotates it when pushed. & Load both small products and the red order card into the brown shipping tray, while it is in the yellow loading rectangle. Then push the loaded tray with the closed gripper completely inside the green delivery rectangle. All three items must remain inside the tray, with the gripper open and clear. \\
Parts Cart & L6 (L4, L2, L5) & A clear panel blocks retrieval from one side; releasing the second part displaces the first; the platform's off-center mass rotates it when pushed. & Use the wooden tool to retrieve the red part from under the low glass table. Load the red and blue parts on the two marks on the brown transport platform, then push the loaded platform with the closed gripper completely into the green parking area. At the end both parts must still be on their marks, and the gripper must be open and clear. \\
Breakfast Service & L6 (L1, L3, L4) & A command offset makes the butter grasp miss; pulling the carton straight back cannot free it from its cradle; the upright butter keeps the lid from closing. & Prepare breakfast: put the bread from the wooden box onto the plate and close the box; pour 80 to 160 mL of milk into the cup using the milk carton in its cradle. Then store the butter inside the box and leave its lid shut. Finish with the carton upright, bread on the plate, cup upright, and gripper open and clear. Spill at most 15 mL. The cradle can be tipped with its gold side handle using ordinary gripper motions. \\
Clear the Desk & L6 (L1, L3, L2) & A command offset makes the first grasp miss; the book cannot be lifted through the tray above it; removing the book drops the tray and undoes the mug placement. & Clear the desk: place the red paperweight, the lidded mug and the yellow supply jar upright on the side tray, and put the black book into the organizer bin. Leave the book fully clear of the tray, all items stable, and the gripper open and clear. The mug and small jars have graspable lid knobs. The book has a side pull tab. The book must end in the organizer, not underneath the tray. \\
Two-Station Packing & L6 (L1, L2, L4) & The command offset differs between the two workstations; a wrong correction disturbs the placed red module; the bar laid across the narrow tray does not fit. & Prepare the two modules for handover: stand the red module on its red square, and fit the entire long blue module inside the narrow tray. Leave both released and stable, with the gripper open and clear. The red module has a graspable top knob. The work areas are on opposite sides of the desk. \\
Wedge Relay & L6 (L1, L4, L2) & A command offset makes the wedge grasp miss; a sideways wedge does not enter the tray slot; removing the wedge lets the first parcel slide off. & Put the red parcel on the right tray and the blue parcel on the left tray. There is one handled support wedge available for the two trays. Finish with both parcels resting entirely on their trays, the wedge returned to its yellow parking area, and the gripper open and clear. Each tray has a shallow retaining pocket and guide rails underneath. \\
\end{longtable}
}

\twocolumn[{\begin{minipage}{\textwidth}
\begingroup\captionsetup{hypcap=false}
\centering
\captionof{table}{\textbf{Persistence of decisive choices across retries.} Main-run retries pool attempts across GPT-6 Astra, Opus 5.5, Sonnet 5.5, and DeepSeek V4.1 Flash under standard binary outcome feedback. The right column evaluates fresh agents explicitly informed of the hidden physical cause.}\label{tab:s2}
\normalsize
\setlength{\tabcolsep}{4pt}
\renewcommand{\arraystretch}{1.12}
\begin{tabular}{@{}
  >{\raggedright\arraybackslash}p{(\linewidth - 6\tabcolsep) * \real{0.1500}}
  >{\raggedright\arraybackslash}p{(\linewidth - 6\tabcolsep) * \real{0.2300}}
  >{\raggedright\arraybackslash}p{(\linewidth - 6\tabcolsep) * \real{0.3900}}
  >{\raggedright\arraybackslash}p{(\linewidth - 6\tabcolsep) * \real{0.2300}}@{}}
\toprule\noalign{}
\begin{minipage}[b]{\linewidth}\raggedright
Task
\end{minipage} & \begin{minipage}[b]{\linewidth}\raggedright
Decisive physical choice
\end{minipage} & \begin{minipage}[b]{\linewidth}\raggedright
Retries in the main runs
\end{minipage} & \begin{minipage}[b]{\linewidth}\raggedright
Told the cause
\end{minipage} \\
\midrule\noalign{}
Clear Cover & Remove the transparent lid & Lid removed in 0\% of attempts (0/36) & Opus succeeds in 75\% (3/4) \\
Card on Shelf & Slide out the unweighted end & After failing with the weighted end, GPT-6 Astra and Opus slide it out again in 91\% (10/11) & Opus and GPT-6 Astra succeed in 100\% (9/9) \\
Tidy Stack & Do not load the empty carton & After a collapse, the next attempt collapses again in 88\% (15/17) & Opus succeeds in 100\% (4/4) \\
Fit Container & Select the box that fits the clear caps & Right box chosen: GPT-6 Astra 50\% (4/8), Opus 22\% (2/9), Sonnet 22\% (2/9), DeepSeek 0\% (0/9) & Opus succeeds in 100\% (6/6) \\
\bottomrule
\end{tabular}
\endgroup
\end{minipage}\vspace{12pt}
}]


\begin{table*}[t]
\centering
\caption{\textbf{Feedback ladder across four cumulative information levels.} Within each model, fresh agents are restarted for one retry from that model's exact same failed first-attempt records across eight resettable {L4}--{L5} tasks ({15} instances for Opus 5.5, {18} for Sonnet 5.5, and {14} for GPT-6 Astra).}\label{tab:s3}
\small
\setlength{\tabcolsep}{4pt}
\renewcommand{\arraystretch}{1.12}
\begin{tabular}{@{}
  >{\raggedright\arraybackslash}p{(\linewidth - 8\tabcolsep) * \real{0.2400}}
  >{\centering\arraybackslash}p{(\linewidth - 8\tabcolsep) * \real{0.1800}}
  >{\centering\arraybackslash}p{(\linewidth - 8\tabcolsep) * \real{0.1800}}
  >{\centering\arraybackslash}p{(\linewidth - 8\tabcolsep) * \real{0.1900}}
  >{\centering\arraybackslash}p{(\linewidth - 8\tabcolsep) * \real{0.2100}}@{}}
\toprule\noalign{}
\begin{minipage}[b]{\linewidth}\raggedright
Model / Round
\end{minipage} & \begin{minipage}[b]{\linewidth}\centering
F1 (Outcome)
\end{minipage} & \begin{minipage}[b]{\linewidth}\centering
F2 (+ Quant. Error)
\end{minipage} & \begin{minipage}[b]{\linewidth}\centering
H+ (+ Cause)
\end{minipage} & \begin{minipage}[b]{\linewidth}\centering
Fix (+ Verified Repair)
\end{minipage} \\
\midrule\noalign{}
\textbf{Opus 5.5} & & & & \\
Round 1 & 5/15 & 8/15 & 14/15 & 12/15 \\
Round 2 & 8/15 & 8/15 & 14/15 & 13/15 \\
\textbf{Pooled (\%)} & \textbf{43.3} & \textbf{53.3} & \textbf{93.3} & \textbf{83.3} \\
\textbf{Sonnet 5.5} & & & & \\
Round 1 & 6/18 & 4/18 & 10/18 & 9/18 \\
Round 2 & 4/18 & 5/18 & 12/18 & 12/18 \\
\textbf{Pooled (\%)} & \textbf{27.8} & \textbf{25.0} & \textbf{61.1} & \textbf{58.3} \\
\textbf{GPT-6 Astra} & & & & \\
Round 1 & 4/14 & 7/14 & 10/14 & 12/14 \\
Round 2 & 4/14 & 7/14 & 10/14 & 12/14 \\
\textbf{Pooled (\%)} & \textbf{28.6} & \textbf{50.0} & \textbf{71.4} & \textbf{85.7} \\
\bottomrule
\end{tabular}
\end{table*}

\section{Detailed Analysis of Failure, Recovery, and Episode Anatomy}\label{b.-detailed-analysis-of-failure-recovery-and-episode-anatomy}

In \hyperref[sec:findings]{Sec.~\ref*{sec:findings}}, we summarized five core findings explaining why certain manipulation failures are repaired autonomously while others persist across attempts. Here we provide the underlying quantitative breakdowns---examining physical choice persistence (\hyperref[b.1.-persistence-of-failed-physical-choices]{Sec. B.1}), progressive causal feedback (\hyperref[b.2.-failure-feedback-ladder-and-causal-diagnosis]{Sec. B.2}), and retained interaction memory (\hyperref[b.3.-recovery-with-retained-interaction-history]{Sec. B.3})---followed by a fine-grained trajectory trace illustrating how causal diagnosis alters physical interaction on the exact same scene instance (\hyperref[b.4.-case-study-anatomy-of-a-failure-recovery-episode]{Sec. B.4}).

\subsection{Persistence of Failed Physical Choices}\label{b.1.-persistence-of-failed-physical-choices}

To test whether multimodal agents spontaneously revise flawed physical assumptions under standard binary trial verdicts, we inspect decisive physical choices across retry attempts on four representative {L4} and {L5} tasks from our main evaluation runs (GPT-6 Astra, Claude Opus 5.5, Claude Sonnet 5.5, and DeepSeek V4.1 Flash). As an upper-bound reference for motor feasibility, we compare these retries against fresh agents initialized on the same instances with the hidden physical cause explicitly stated in the prompt. \textbf{\hyperref[tab:s2]{Table S2}} summarizes the results.

Without external causal hints, agents overwhelmingly repeat the exact physical decision that triggered the initial failure. Visually subtle geometric traps remain largely unaddressed: across 36 total attempts on \emph{Clear Cover}, no agent ever removes the transparent lid ({0/36}), and most attempts on \emph{Fit Container} repeatedly select the visually plausible but undersized box. Crucially, persistence is not limited to visually ambiguous traps. Even when failure produces an unmistakable visual outcome---such as the weighted card tipping off the shelf in \emph{Card on Shelf} or the empty carton buckling under load in \emph{Tidy Stack}---frontier agents repeat the same failure-inducing choice in roughly {90\%} of subsequent retries ({10/11} and {15/17}), typically adjusting approach speed or small positional offsets instead of changing which end they slide or which box supports the stack. By contrast, fresh agents informed of the hidden cause succeed in {75\%}--{100\%} of runs, confirming that low-level motor execution is not the limiting factor.

\subsection{Failure Feedback Ladder and Causal Diagnosis}\label{b.2.-failure-feedback-ladder-and-causal-diagnosis}

To separate failure diagnosis from repair execution (Findings 1 and 2, Fig. 4 of the main paper), we construct a four-level feedback ladder across eight representative resettable {L4}--{L5} tasks. Because models fail on different subsets of instances during their initial try, each model is evaluated starting from \textbf{its own failed first-attempt records} drawn from the main evaluation run, yielding {15} failed instances for Opus 5.5, {18} for Sonnet 5.5, and {14} for GPT-6 Astra. Crucially, \textbf{within each model}, all four feedback levels restart fresh agents from the exact same set of failed first-attempt records and allow a single retry, ensuring a strictly paired comparison across feedback conditions:

\begin{enumerate}
\def\labelenumi{\arabic{enumi}.}
\tightlist
\item
  \textbf{Binary outcome ({F1}):} notifies the agent only that the prior attempt failed.
\item
  \textbf{Quantitative error ({F2}):} augments {F1} with the evaluator's measured goal residual (e.g., object-to-target distance or unsatisfied predicate count) without naming the trap.
\item
  \textbf{Hidden physical cause ({H+}):} augments {F2} with a concise statement of the underlying physical obstacle, without specifying how to manipulate the scene.
\item
  \textbf{Verified repair ({Fix}):} augments {H+} with a concrete, simulation-verified action strategy.
\end{enumerate}

\newpage
\textbf{\hyperref[tab:s3]{Table S3}} reports per-round and pooled recovery rates across two independent rounds per model. Two key patterns emerge:

\begin{itemize}
\item
  \textbf{Causal diagnosis is the primary bottleneck for frontier agents.} Moving from binary feedback ({F1}) or quantitative error ({F2}) to causal disclosure ({H+}) produces the largest jump across every model---lifting pooled recovery from {43.3\%} ({F1}) and {53.3\%} ({F2}) to {93.3\%} ({H+}) for Opus 5.5, from {27.8\%} to {61.1\%} for Sonnet 5.5, and from {28.6\%} to {71.4\%} in both rounds and pooled for GPT-6 Astra. Meanwhile, the persistent gap under {H+} between Opus 5.5 ({93.3\%}) and Sonnet 5.5 ({61.1\%}) reflects coupled motor-execution errors in weaker models (Finding 2).
\item
  \textbf{Over-specifying the repair ({Fix}) can induce closed-loop rigidity.} Notably, adding an explicit repair description ({Fix}) slightly \emph{reduces} pooled recovery relative to {H+} for Opus 5.5 ({93.3\%} to {83.3\%}) and Sonnet 5.5 ({61.1\%} to {58.3\%}). Inspection of rollout logs reveals that when given only the physical cause ({H+}), capable agents synthesize their own recovery plan and actively adjust to intermediate visual and contact feedback. By contrast, when handed a prescribed repair maneuver ({Fix}), agents tend to follow the described action sequence mechanically; when minor contact or pose deviations arise mid-rollout, rigid adherence to the suggested trajectory sacrifices closed-loop adaptability. Conversely, GPT-6 Astra---which exhibits lower initial grasp precision and higher slip frequency---gains modestly from the concrete geometric guidance in {Fix} ({71.4\%} to {85.7\%} pooled).
\end{itemize}

\begin{figure*}[t]
\centering
\includegraphics[width=1\linewidth,height=\textheight,keepaspectratio]{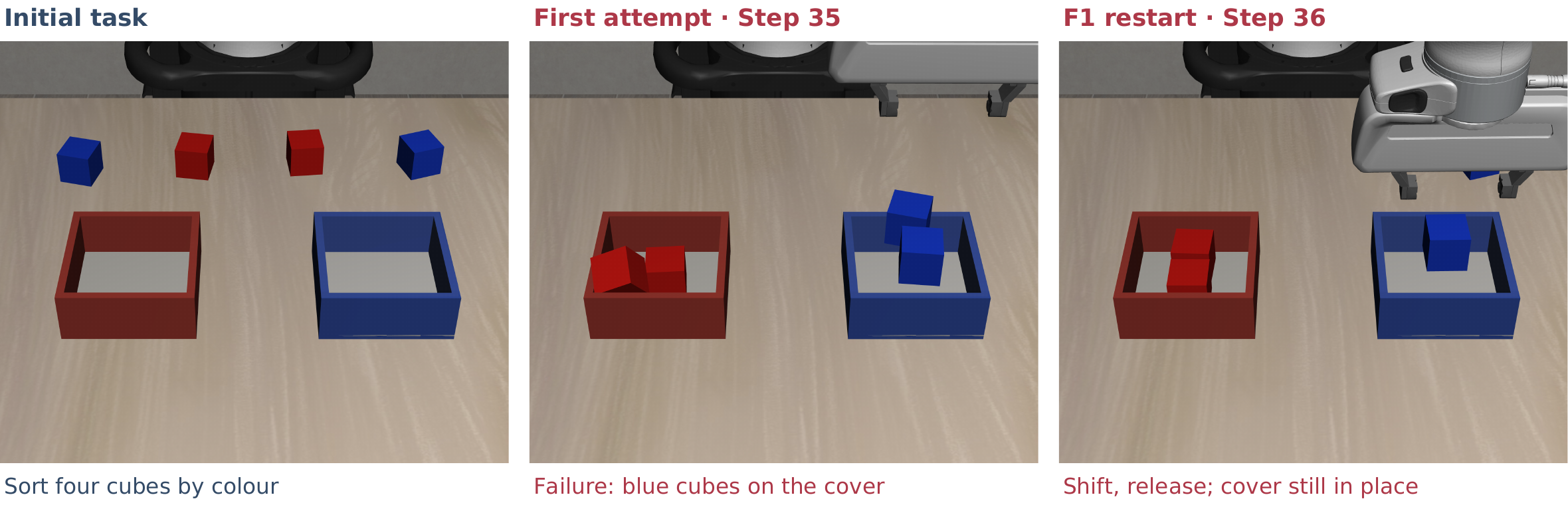}
\caption{\textbf{Failure and {F1} retry on the same \emph{Clear Cover} instance ({seed\ 3101}).} \emph{Left:} Initial scene (\emph{Sort four cubes by colour}). \emph{Middle:} First-attempt outcome at step 35, where the agent claims success while both blue cubes rest on the transparent cover. \emph{Right:} {F1} restart at step 36 after shifting horizontal placement coordinates and releasing; the lid remains in place.}\label{fig:s2}
\end{figure*}

\subsection{Recovery with Retained Interaction History}\label{b.3.-recovery-with-retained-interaction-history}

To determine which components of an agent's prior experience drive cross-attempt adaptation without external causal hints (Finding 5 of the main paper), we examine all {102} instances of the {34} resettable tasks ({L1}, four resetting {L2} tasks, {L3}--{L6}). Conditioning on the {57} instances where a shared first attempt by Opus 5.5 fails ({45/102\ =\ 44.1\%} first-attempt success), we compare four retry configurations over up to two additional attempts:

\textbf{\hyperref[tab:s4]{Table S4}} reports cumulative recoveries within two and three total attempts.

\begin{table}[!htbp]
\centering
\caption{\textbf{Ablation of retained interaction history on Opus 5.5.} Cumulative recoveries across the {57} resettable instances whose shared first attempt failed. All retry conditions, including Full History and record-based restarts (Observation only, Images only, and Memory-free), were evaluated using the same Claude Code release, with identical MCP tool schemas, image resolutions, and controller interfaces.}\label{tab:s4}
\small
\setlength{\tabcolsep}{4pt}
\renewcommand{\arraystretch}{1.12}
\begin{tabular}{@{}
  >{\raggedright\arraybackslash}p{(\linewidth - 6\tabcolsep) * \real{0.2500}}
  >{\centering\arraybackslash}p{(\linewidth - 6\tabcolsep) * \real{0.2300}}
  >{\centering\arraybackslash}p{(\linewidth - 6\tabcolsep) * \real{0.2300}}
  >{\centering\arraybackslash}p{(\linewidth - 6\tabcolsep) * \real{0.2900}}@{}}
\toprule\noalign{}
\begin{minipage}[b]{\linewidth}\raggedright
Retry condition
\end{minipage} & \begin{minipage}[b]{\linewidth}\centering
Recovered within 2 attempts
\end{minipage} & \begin{minipage}[b]{\linewidth}\centering
Recovered within 3 attempts
\end{minipage} & \begin{minipage}[b]{\linewidth}\centering
Overall \(A_3\) across 102 resettable instances (\%)
\end{minipage} \\
\midrule\noalign{}
\textbf{Full history} & \textbf{24/57} & \textbf{36/57} & \textbf{79.4} \\
Observation only & 20/57 & 29/57 & 72.5 \\
Images only & 13/57 & 22/57 & 65.7 \\
Memory-free & 11/57 & 14/57 & 57.8 \\
\bottomrule
\end{tabular}
\end{table}

\begin{enumerate}
\def\labelenumi{\arabic{enumi}.}
\tightlist
\item
  \textbf{Full history:} continues the live multi-turn session from the main evaluation run, preserving all prior reasoning, commands, workspace scripts/notes, robot replies, images, and attempt verdicts.
\item
  \textbf{Observation only:} initializes a fresh agent per retry from a filtered interaction log containing only prior robot state/status replies, camera images, and attempt verdicts (stripping prior agent commands, reasoning traces, and workspace notes).
\item
  \textbf{Images only:} initializes a fresh agent per retry given only prior camera images and attempt verdicts.
\item
  \textbf{Memory-free:} initializes a completely fresh agent per retry with no record of prior attempts.
\end{enumerate}

Recovery rises monotonically with the richness of the retained record: within three attempts, solved failed instances climb from {14/57} ({Memory-free}, corresponding to {59/102\ =\ 57.8\%} overall success) to {22/57} ({Images\ only}), {29/57} ({Observation\ only}), and {36/57} ({Full\ history}, corresponding to {81/102\ =\ 79.4\%} overall success, a {+21.6} percentage-point history gain). Paired across instances using an exact sign test with Holm multiple-testing correction \cite{holm1979simple} over all six condition pairs, both {Full\ history} ({23} wins vs.~{1} loss, \(p < 0.001\)) and {Observation\ only} ({17} wins vs.~{2} losses, \(p = 0.004\)) significantly outperform {Memory-free} retries, and {Full\ history} also significantly outperforms {Images\ only} ({17} wins vs.~{3} losses, \(p = 0.01\)). Thus, quantitative proprioceptive/motion replies ({Observation\ only}) and explicit action-reasoning context ({Full\ history}) carry critical diagnostic signal that passive visual frames alone ({Images\ only}) fail to capture. Even with full history retained, however, {21/57} instances remain unsolved after three attempts, motivating structured probing and choice-revision strategies (\hyperref[c.-from-failure-analysis-to-recovery-strategies]{Sec. C}).

\begin{figure*}[t]
\centering
\includegraphics[width=1\linewidth,height=\textheight,keepaspectratio]{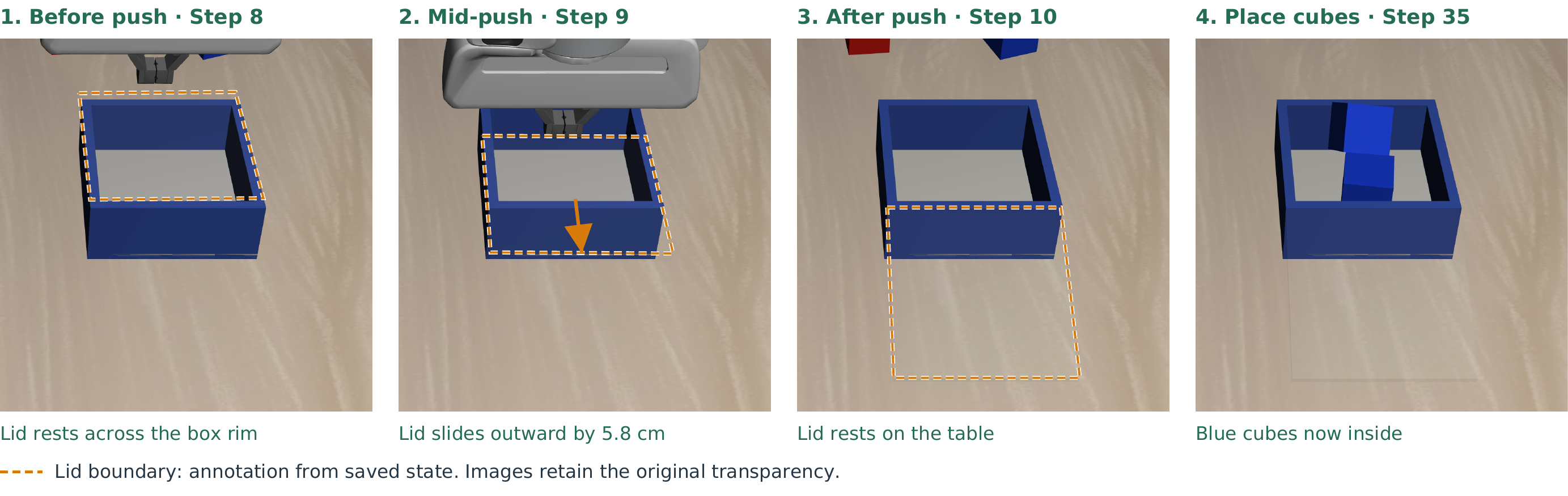}
\caption{\textbf{Causal diagnosis changes physical interaction.} Following an empty-gripper vertical probe, the cause-informed agent sweeps the transparent lid off the blue box and places both blue cubes inside: (1) contact preparation before the push (step 8), (2) mid-push as the lid slides outward by \(5.8\text{ cm}\) (step 9), (3) lid resting clear on the table (step 10), and (4) verified cube placement inside the cleared box (step 35). Orange dashed outlines annotate saved lid geometry for the reader; images rendered to the agent retain original transparency.}\label{fig:s3}
\end{figure*}

\subsection{Case Study: Anatomy of a Failure-Recovery Episode}\label{b.4.-case-study-anatomy-of-a-failure-recovery-episode}

To illustrate concretely how anomaly rationalization causes retries to stall---and how causal insight transforms physical interaction---we trace Opus 5.5 across three runs on the exact same \emph{Clear Cover} instance ({seed\ 3101}): (1) the initial attempt, (2) a standard {F1} retry retaining interaction history, and (3) a cause-informed ({H+}) restart. The task requires sorting two red cubes into the red box and two blue cubes into the blue box, where a nearly invisible transparent lid rests across the rim of the blue box (stopping downward motion at end-effector height \(z = 0.0614\text{ m}\), roughly \(2\text{ cm}\) above the interior placement height).

\Needspace{5\baselineskip}
\textbf{1. First attempt: localize in 2D, place blindly, and rationalize completion.} Inspecting the initial scene (\textbf{\hyperref[fig:s2]{Figure S2}}, left), the agent writes a workspace script ({seg.py}) to segment red and blue pixels from the wrist camera view and compute 2D grasp centroids:

\begin{Verbatim}[fontsize=\footnotesize,breaklines=true,breakanywhere=true]
import cv2, numpy as np, sys
im = cv2.imread(sys.argv[1])
b, g, r = [im[:, :, i].astype(int) for i in range(3)]
for name, m in [('red', (r > 120) & (g < 60) & (b < 60)),
                ('blue', (b > 120) & (r < 60) & (g < 80))]:
    n, l, st, c = cv2.connectedComponentsWithStats(m.astype(np.uint8))
    for i in range(1, n):
        if st[i, 4] > 30:
            print(name, st[i].tolist(), c[i].round(1).tolist())
\end{Verbatim}

On the wrist view ({s002\_wrist.jpg}), this script accurately localizes the pixel centroids ({red\ {[}206.6,\ 366.8{]}}, {blue\ {[}107.3,\ 363.7{]}}), guiding a clean grasp at step 4 ({move\_eef\ x=0.34\ y=0.066\ z=0.2\ yaw=0}). Because 2D color thresholding is completely blind to the clear lid, the agent executes standard pick-and-place motions over both boxes and calls {finish} claiming task completion. In reality, while both red cubes enter the red box, both blue cubes come to rest atop the transparent cover (\textbf{\hyperref[fig:s2]{Figure S2}}, middle).

\Needspace{5\baselineskip}
\textbf{2. {F1} restart: notice the contact anomaly, misattribute the cause, and tweak coordinates.} Restarted from the failed first-attempt record with binary failure feedback ({F1}), a fresh agent calibrates an explicit camera-to-base projection model and verifies the horizontal bounds of the blue box. During blue-cube placement (step 31), its commanded descent to \(z = 0.045\text{ m}\) arrests early at \(z = 0.0614\text{ m}\). Rather than questioning whether the box opening is obstructed, the agent attributes this premature contact to grazing the front box wall. It lifts the cube, shifts its horizontal target inward, and commands another descent (steps 32--34), stalling once again at \(z = 0.0614\text{ m}\). Upon opening the gripper and inspecting the top-down view, it rationalizes that the cube is resting on the box floor and spends its remaining tool-call budget making minor \((x, y)\) adjustments with the lid still untouched (\textbf{\hyperref[fig:s2]{Figure S2}}, right).

\Needspace{5\baselineskip}
\textbf{3. Cause-informed ({H+}) restart: actively probe support and slide off the obstruction.} A separate fresh agent receives the exact same failed first-attempt record augmented only with the physical cause (providing zero action coordinates):

\begin{quote}
\emph{``A cube released over that box stops on the lid at rim height; it is not in the box.''}
\end{quote}

Starting from the reset initial scene, the agent immediately forms a physical test: it closes the empty gripper above the blue box and commands a slow downward probe ({speed=0.03}) toward \(z = 0.040\text{ m}\), which arrests at \(z = 0.0614\text{ m}\)---confirming an obstruction roughly \(2.1\text{ cm}\) above its target (step 5). Using this measured rim height, it positions the closed gripper just below the lid edge at the near rim (\(x = 0.465\text{ m}, z = 0.057\text{ m}\)) and executes a controlled horizontal sweep to \(x = 0.670\text{ m}\) (steps 7--9), sliding the transparent lid cleanly off the box onto the table (\textbf{\hyperref[fig:s3]{Figure S3}}). It then places the first blue cube (step 28), visually confirms that the cube dropped inside the box interior, places the remaining cubes, and completes all five evaluator predicates in 35 motion commands.

\section{From Failure Analysis to Recovery Strategies}\label{c.-from-failure-analysis-to-recovery-strategies}

In \hyperref[sec:harness]{Secs.~\ref*{sec:harness}} and \hyperref[sec:real_world]{\ref*{sec:real_world}}, we compared external robot harnesses against our minimalist prompt-only harness, \textbf{LEAF-Lite}, under a strictly unified MCP protocol. Below we document the controlled porting protocol for external robot harnesses (\hyperref[c.1.-controlled-ports-of-external-robot-harnesses]{Sec. C.1}) and present the exact prompt and design mapping of LEAF-Lite (\hyperref[c.2.-design-and-verbatim-prompt-of-leaf-lite]{Sec. C.2}).

\begin{table*}[t]
\centering
\caption{\textbf{Controlled adaptations of external robot harnesses.} All ports preserve each framework's core planning, reflection, and within-episode memory logic while standardizing physical execution and termination over the unified MCP interface.}\label{tab:s6}
\small
\setlength{\tabcolsep}{4pt}
\renewcommand{\arraystretch}{1.12}
\begin{tabular}{@{}
  >{\raggedright\arraybackslash}p{(\linewidth - 4\tabcolsep) * \real{0.2000}}
  >{\raggedright\arraybackslash}p{(\linewidth - 4\tabcolsep) * \real{0.4000}}
  >{\raggedright\arraybackslash}p{(\linewidth - 4\tabcolsep) * \real{0.4000}}@{}}
\toprule\noalign{}
\begin{minipage}[b]{\linewidth}\raggedright
Robot harness
\end{minipage} & \begin{minipage}[b]{\linewidth}\raggedright
Retained orchestration and memory mechanisms
\end{minipage} & \begin{minipage}[b]{\linewidth}\raggedright
Interface adaptations to LEAF's MCP protocol
\end{minipage} \\
\midrule\noalign{}
\textbf{RPent} \cite{zhang2026harnessvla} & Native Codex planner; multi-session handoff loop that starts a fresh session with a written handoff summary when an attempt ends unfinished (1 attempt per session, outer loop cap set to 5 sessions); task-agnostic prompt rules requiring in-place recovery first, changing a named decisive choice on each retry, archiving failed attempts, and keeping workspace notes. & Custom robot/environment services replaced by the three MCP tools ({robot}, {view\_images}, {shell}); episode completion governed strictly by the benchmark server's {finished} flag (enforcing the standard 3-attempt episode cap); handoff summaries formatted over MCP observations and verdicts. \\
\textbf{SenseNova-RoboRSI} \cite{sensenova2026embodied} & Native Codex planner, event logging, and full conversation context retention, with all attempts of an episode continuing inside a single session. & Original system prompt describing custom action formats replaced with three task-agnostic interface sentences; adapter performs pure schema translation for MCP tool calls and returned results. \\
\textbf{EmbodiedEvo} \cite{li2026embodiedevo} & Native launch and session-continuation command construction; iterative workflow and skill/lesson logging protocol extracted verbatim from its {playground.md} specification, initialized from an empty directory per episode and accumulated strictly within the current episode. & Action plans may still be computed in {shell}, but every physical primitive must execute step-by-step through the MCP {robot} tool rather than direct simulator or scripted skill calls. \\
\bottomrule
\end{tabular}
\end{table*}

\subsection{Controlled Ports of External Robot Harnesses}\label{c.1.-controlled-ports-of-external-robot-harnesses}

To isolate high-level orchestration design from differences in foundation models, observation spaces, or low-level control primitives, all external robot harnesses in \hyperref[tab:harness]{Table~\ref*{tab:harness}}---RPent \cite{zhang2026harnessvla}, SenseNova-RoboRSI \cite{sensenova2026embodied}, and EmbodiedEvo \cite{li2026embodiedevo}---are ported onto a strictly identical runtime and MCP boundary matching the direct Codex baseline:

\begin{itemize}
\item
  \textbf{Identical MCP boundary and model runtime.} Every harness receives only the initial task instruction and connects to the same MCP server exposing {robot} (state query, motion execution, and attempt termination), {view\_images} (visual inspection), and {shell} (sandboxed computation and workspace file I/O). Server-provided tool descriptions and parameter schemas are forwarded verbatim without rewriting. All methods run {gpt-6.1-sol} at {high} reasoning effort on {Codex\ 0.163.0} with identical model directories, disabled built-in features, and read-only sandbox restrictions outside the episode workspace.
\item
  \textbf{Evaluator-owned lifecycle and strict isolation.} Per-attempt tool-call counts, simulated motion-time limits, attempt transitions, scene resets, and outcome verdicts are governed exclusively by the benchmark server. Harnesses cannot trigger resets, judge task completion, access the simulator, or inspect evaluator logs and task code; the model's {shell} runs inside an isolated container with no read access to the harness's own source directory.
\item
  \textbf{Automated preflight and mock-model verification.} Before every run, an automated preflight check strips the harness-specific prompt prefix from the first outgoing model request and verifies an exact field-by-field match against the direct Codex baseline across tool schemas, developer instructions, sampling parameters, and sandbox settings, aborting on any discrepancy. Prior to evaluation, replaying a deterministic scripted mock model through each harness is verified to produce event logs identical to the direct baseline, alongside bit-exact task text and image-byte delivery, clean workspace initialization, and proper cross-attempt continuation.
\end{itemize}

\textbf{\hyperref[tab:s6]{Table S5}} summarizes the orchestration mechanisms retained from each framework and the interface adaptations required to connect it to our MCP server.

\subsection{Design and Verbatim Prompt of LEAF-Lite}\label{c.2.-design-and-verbatim-prompt-of-leaf-lite}

Rather than introducing multi-agent wrappers or external memory pipelines, LEAF-Lite translates our empirical failure findings (\hyperref[b.-detailed-analysis-of-failure-recovery-and-episode-anatomy]{Sec. B}) directly into five task-agnostic behavioral rules appended to the agent harness's developer instructions. No task identities, object geometries, or trap categories are mentioned, preserving identical tools, budgets, and single-session-per-instance execution as the direct baseline:

\begin{itemize}
\item
  \textbf{Active physical probing (Rules 1 and 5 \(\leftarrow\) Findings 1 and 4):} Prompts small, reversible diagnostic probes before committing to irreversible motions and forbids rationalizing premature stalls or unexpected contact anomalies without testing the underlying surface assumption.
\item
  \textbf{Cumulative subgoal verification (Rules 2 and 3 \(\leftarrow\) Findings 3 and 4):} Requires a fresh post-release observation after every placement, cumulative re-verification of all previously completed subgoals to catch cross-stage regressions, and evidence-backed goal verification prior to calling {finish}.
\item
  \textbf{Factual logging and named choice revision (Rule 4 \(\leftarrow\) Findings 1 and 5):} Adapting factual note-taking and lever-level revision from RPent \cite{zhang2026harnessvla}, requires recording concise observations in {notes.md} after any failed attempt and altering at least one named structural physical choice (e.g., object, sequence order, grasp/push point, or direction) rather than merely tuning speed, force, or minor offsets.
\end{itemize}

The exact prompt text appended by LEAF-Lite is reproduced verbatim below:

\begin{quote}
\small

\textbf{Working method for physical manipulation}\\*
The environment does not tell you why an attempt failed. Follow these rules.

\textbf{1. Probe before anything irreversible}\\*
Before an action you cannot easily undo (releasing an object, a fast push or pull, putting one object on top of another, pushing toward an edge), spend one or two commands on a small, reversible test of the assumption that action depends on, and look at the result: for example, lower slowly with the gripper closed, push a short distance, or lift partway. A probe must itself follow the task rules: if the task allows an action only once, the probe must not be that action.

\textbf{2. Observe after every release, check every condition before finish}\\*
After every placement or release, observe once more and check that the object is where it should be and has stayed there. Before {finish}, check each goal condition of the instruction one by one against a fresh observation. Claim success only when you can point to evidence; if something is off and budget remains, fix it in this attempt.

\textbf{3. Keep a subgoal chain}\\*
Split the instruction into ordered subgoals. Each time you complete one, re-check every subgoal completed so far against a fresh observation (after subgoal 3, check subgoals 1, 2 and 3). If an earlier one has been undone, restore it before moving on.

\textbf{4. After a failed attempt, write facts and change a named choice}\\*
Before acting again, write a short factual note to {notes.md} in the workspace: what you did (key commands and numbers), what you observed, and the decisive choices you made. The next attempt must change at least one decisive choice, named in the note (which object, order, grasp point, push point, direction, distance), not just speed, force or a small offset. Do not return to a choice that already failed.

\textbf{5. Do not explain anomalies away}\\*
If a motion stops short of its target, the gripper closes on nothing, or an object moves in a way you did not intend, some belief of yours about the scene is wrong. Check it with a probe (rule 1) before continuing.
\end{quote}

\section{\texorpdfstring{Statistical Reliability and\\Confidence Intervals}{Statistical Reliability and Confidence Intervals}}\label{d.-statistical-reliability}

Each of LEAF's 40 tasks has three parametric instances, giving 120 instances per model. We report first-attempt success \(A_1\) and cumulative success within three attempts \(A_3\). As detailed in \hyperref[a.-benchmark-task-specifications-and-metric-definitions]{Sec. A}, the six {L2} tasks marked with \textsuperscript{$\dagger$} in \hyperref[tab:s1]{Table S1} continue from the disturbed scene state after failure; the other {34} tasks reset to their initial state before each retry.

\textbf{\hyperref[tab:s7]{Table S6}} reports {95\%} Wilson score confidence intervals \cite{wilson1927probable} across all {120} instances for the four representative models spanning the capability spectrum. To account for within-task correlation across the three seeds of each task, we evaluate pairwise model differences using two-sided paired permutation tests clustered at the task level with Holm step-down correction \cite{holm1979simple}:

\begin{itemize}
\item
  \textbf{Frontier tier:} There is no statistically significant difference between GPT-6 Astra and Claude Opus 5.5 on either \(A_1\) ({50.0\%} vs.~{47.5\%}) or \(A_3\) ({86.7\%} vs.~{82.5\%}).
\item
  \textbf{Cross-tier separation:} Opus 5.5 significantly outperforms Sonnet 5.5 on both \(A_1\) (\(p = 0.001\)) and \(A_3\) (\(p < 0.001\)), and Sonnet 5.5 significantly outperforms DeepSeek V4.1 Flash on both \(A_1\) (\(p = 0.005\)) and \(A_3\) (\(p < 0.001\)).
\item
  \textbf{Run-to-run repeatability:} Across the repeated evaluation rounds of the feedback ladder (\hyperref[tab:s3]{Table S3}), running a fresh agent on the exact same failed instance at the same feedback level yields identical binary outcomes in {54/60} ({90.0\%}) paired runs for Opus 5.5 and {52/72} ({72.2\%}) for Sonnet 5.5, with similarly stable aggregate recovery across both rounds of GPT-6 Astra, confirming high evaluation reproducibility.
\end{itemize}

\textbf{Paired harness comparison.} For the \(A_3\) results in \hyperref[tab:harness]{Table~\ref*{tab:harness}}, LEAF-Lite solves 108/120 instances and RPent solves 101/120, a difference of 5.8 percentage points. Among the paired outcomes, eight instances are solved only by LEAF-Lite and one only by RPent. A two-sided exact McNemar test gives unadjusted \(p=0.0391\). This instance-level analysis does not account for dependence among the three instances of each task.

\begin{table}[t]
\centering
\caption{\textbf{Overall success rates and 95\% Wilson confidence intervals.} Evaluated across all {120} instances ({40} tasks \(\times\) {3} seeds) without an external robot harness; {95\%} Wilson intervals \cite{wilson1927probable} are shown in brackets.}\label{tab:s7}
\small
\setlength{\tabcolsep}{4pt}
\renewcommand{\arraystretch}{1.12}
\begin{tabular}{@{}
  >{\raggedright\arraybackslash}p{(\linewidth - 6\tabcolsep) * \real{0.2200}}
  >{\raggedright\arraybackslash}p{(\linewidth - 6\tabcolsep) * \real{0.1800}}
  >{\centering\arraybackslash}p{(\linewidth - 6\tabcolsep) * \real{0.3000}}
  >{\centering\arraybackslash}p{(\linewidth - 6\tabcolsep) * \real{0.3000}}@{}}
\toprule\noalign{}
\begin{minipage}[b]{\linewidth}\raggedright
Agent
\end{minipage} & \begin{minipage}[b]{\linewidth}\raggedright
Agent harness
\end{minipage} & \begin{minipage}[b]{\linewidth}\centering
\(A_1\) (\%) {[}95\% CI{]}
\end{minipage} & \begin{minipage}[b]{\linewidth}\centering
\(A_3\) (\%) {[}95\% CI{]}
\end{minipage} \\
\midrule\noalign{}
GPT-6 Astra & Codex & 50.0 {[}41.2, 58.8{]} & 86.7 {[}79.4, 91.6{]} \\
Opus 5.5 & Claude Code & 47.5 {[}38.8, 56.4{]} & 82.5 {[}74.7, 88.3{]} \\
Sonnet 5.5 & Claude Code & 26.7 {[}19.6, 35.2{]} & 54.2 {[}45.3, 62.8{]} \\
DeepSeek V4.1 Flash & Codex & 7.5 {[}4.0, 13.6{]} & 24.2 {[}17.4, 32.6{]} \\
\bottomrule
\end{tabular}
\end{table}

\begin{table*}[t]
\centering
\caption{\textbf{Fine-tuning configurations for VLA and WAM policies.} All policies are trained in {bf16} across {16} PPUs with approximately {95.6\ GiB} of device memory each. \emph{Action chunk (executed)} denotes the predicted horizon vs.~the number of steps executed open-loop before re-querying the policy.}\label{tab:s8}
\small
\setlength{\tabcolsep}{4pt}
\renewcommand{\arraystretch}{1.12}
\begin{tabular}{@{}
  >{\raggedright\arraybackslash}p{(\linewidth - 8\tabcolsep) * \real{0.1800}}
  >{\raggedright\arraybackslash}p{(\linewidth - 8\tabcolsep) * \real{0.1700}}
  >{\raggedright\arraybackslash}p{(\linewidth - 8\tabcolsep) * \real{0.2300}}
  >{\raggedright\arraybackslash}p{(\linewidth - 8\tabcolsep) * \real{0.2100}}
  >{\raggedright\arraybackslash}p{(\linewidth - 8\tabcolsep) * \real{0.2100}}@{}}
\toprule\noalign{}
\begin{minipage}[b]{\linewidth}\raggedright
Hyperparameter
\end{minipage} & \begin{minipage}[b]{\linewidth}\raggedright
\(\pi_{0.5}\) \cite{physicalintelligence2025pi05}
\end{minipage} & \begin{minipage}[b]{\linewidth}\raggedright
DM0.5-MEM \cite{dexmal2026dm05}
\end{minipage} & \begin{minipage}[b]{\linewidth}\raggedright
FastWAM \cite{yuan2026fastwam}
\end{minipage} & \begin{minipage}[b]{\linewidth}\raggedright
OpenWAM-\(\alpha\) \cite{wang2026openwam}
\end{minipage} \\
\midrule\noalign{}
Initialization & \(\pi_{0.5}\) base & DM0.5-MEM pretrained & Wan2.2 TI2V-5B \cite{wan2025wan} video model + action expert & OpenWAM-\(\alpha\) pretrained \\
Image input & 2 views, \(224 \times 224\) each & 2 views at \(448\text{ px}\); 32 earlier external frames at \(1\text{ Hz}\) from current attempt & 2 views tiled, \(224 \times 448\) & 2 views tiled, \(384 \times 320\) \\
Proprioceptive input & None & None & Current 8-D state & Current 8-D state \\
Action chunk (executed) & 10 (5) & 10 (10) & 32 (10) & 32 (10) \\
Optimizer & AdamW \cite{loshchilov2019decoupled} & MuonAdamW \cite{jordan2024muon} & AdamW \cite{loshchilov2019decoupled} & AdamW \cite{loshchilov2019decoupled} \\
Learning rate schedule & \(5 \times 10^{-5}\), 10k warmup steps, then constant & \(2.5 \times 10^{-5}\) peak, cosine decay to \(0.1\times\) & \(10^{-4}\), 5\% warmup, cosine decay & \(10^{-4}\), 5\% warmup, cosine; continued at \(5 \times 10^{-5}\) \\
Global batch size & 256 & 128, then 256 & 256 & 256 \\
Checkpoint in Tab. 1 & 50k steps & 9,213 steps (6.0 epochs) & 12k steps (6.7 epochs) & 12k + 8.6k steps (12.0 epochs) \\
\bottomrule
\end{tabular}
\end{table*}

\section{Learned Action Policies: Implementation and Ablations}\label{e.-learned-action-policies-implementation-and-ablations}

Complementing \hyperref[sec:setup]{Secs.~\ref*{sec:setup}} and \hyperref[sec:main]{\ref*{sec:main}} and \hyperref[tab:main]{Table~\ref*{tab:main}}, \hyperref[e.1.-policy-fine-tuning-and-low-level-evaluation-interface]{Sec. E.1} details fine-tuning and low-level control for the Vision--Language--Action (VLA) and World Action Model (WAM) baselines: \(\pi_{0.5}\) \cite{physicalintelligence2025pi05}, DM0.5-MEM \cite{dexmal2026dm05}, FastWAM \cite{yuan2026fastwam}, and OpenWAM-\(\alpha\) \cite{wang2026openwam}. \hyperref[e.2.-training-with-outcome-labeled-failure-rollouts]{Sec. E.2} examines training with outcome-labeled failures, and \hyperref[e.3.-role-of-within-attempt-temporal-context-at-inference]{Sec. E.3} examines visual frame history at inference.

\subsection{Policy Fine-Tuning and Low-Level Evaluation Interface}\label{e.1.-policy-fine-tuning-and-low-level-evaluation-interface}

\textbf{Training data.} All VLA and WAM baselines in \hyperref[tab:main]{Table~\ref*{tab:main}} are fine-tuned on the exact same curated set of {511} successful demonstration trajectories ({457,583} control frames at {20\ Hz}) drawn from our replay dataset (\hyperref[f.-replay-dataset-documentation]{Sec. F}). Demonstrations come exclusively from the non-held-out training instances of the single-cause tasks; because \emph{Clear Cover} and \emph{Tidy Stack} exhibit sparse successes on the two default training seeds, we include successful rollouts from four additional non-held-out random seeds for those two tasks, while strictly holding out all evaluation instances ({seed\ 3} of every task). Post-attempt settling frames are trimmed prior to training. All models condition on the natural-language task instruction and predict {7-D} action chunks (normalized end-effector pose deltas plus gripper command).

\textbf{Compute and hyperparameters.} \textbf{\hyperref[tab:s8]{Table S7}} summarizes the model initializations, visual/state inputs, action-chunk horizons, and optimization schedules. All models are fine-tuned in bf16 precision using 16 parallel PPU-ZW810E accelerators, each equipped with approximately 95.6 GiB of device memory.

\textbf{Matched low-level control interface.} The simulation environment streams two {256\ ×\ 256} RGB camera views from {agentview} and the wrist, the {8-D} proprioceptive state (end-effector Cartesian position, axis-angle orientation, and two gripper finger positions), and the verbatim task instruction; each policy resizes or tiles its visual inputs according to Table S7. Predicted {7-D} actions are executed directly by the same {20\ Hz} operational-space controller that underlies the high-level {move\_eef} primitive used by agentic models. Each attempt is allocated the identical simulated-time budget as agent-as-policy runs and terminates early if the arm remains stationary for {5\ s}. Every instance is evaluated over up to three attempts under the standard benchmark reset rules without external feedback. DM0.5-MEM clears its temporal frame buffer across resets. Evaluation covers the {31} held-out {L2}--{L5} instances. {L1} tasks are omitted because their camera-gripper command offsets are injected inside the high-level {move\_eef} tool, which direct {20\ Hz} low-level actions bypass.

\begin{table*}[t]
\centering
\caption{\textbf{Learned-policy ablations on failure-data conditioning and temporal context.} Number of solved instances on the {31} held-out {L2}--{L5} evaluation instances (matching \hyperref[tab:main]{Table~\ref*{tab:main}}) and the {62} {L2}--{L5} training instances (disjoint random seeds). Step counts denote optimization steps.}\label{tab:s9}
\small
\setlength{\tabcolsep}{4pt}
\renewcommand{\arraystretch}{1.12}
\begin{tabular}{@{}
  >{\raggedright\arraybackslash}p{(\linewidth - 8\tabcolsep) * \real{0.3600}}
  >{\centering\arraybackslash}p{(\linewidth - 8\tabcolsep) * \real{0.1600}}
  >{\centering\arraybackslash}p{(\linewidth - 8\tabcolsep) * \real{0.1600}}
  >{\centering\arraybackslash}p{(\linewidth - 8\tabcolsep) * \real{0.1600}}
  >{\centering\arraybackslash}p{(\linewidth - 8\tabcolsep) * \real{0.1600}}@{}}
\toprule\noalign{}
\begin{minipage}[b]{\linewidth}\raggedright
Policy / Variant
\end{minipage} & \begin{minipage}[b]{\linewidth}\centering
Held-out (31) \(A_1\)
\end{minipage} & \begin{minipage}[b]{\linewidth}\centering
Held-out (31) \(A_3\)
\end{minipage} & \begin{minipage}[b]{\linewidth}\centering
Training (62) \(A_1\)
\end{minipage} & \begin{minipage}[b]{\linewidth}\centering
Training (62) \(A_3\)
\end{minipage} \\
\midrule\noalign{}
\textbf{Model: \(\pi_{0.5}\)} & & & & \\
Success only, 30k & 2 & 5 & 9 & 14 \\
Success + failure, 30k & 3 & 3 & 2 & 5 \\
Success only, 50k \emph{(Tab. 1)} & 2 & 7 & 6 & 13 \\
Success + failure, 50k & 1 & 2 & 4 & 10 \\
\textbf{Model: DM0.5-MEM} & & & & \\
With history \emph{(Tab. 1)} & 5 & 6 & 11 & 14 \\
Success + failure & 0 & 1 & 3 & 6 \\
History removed at inference & 0 & 3 & 1 & 4 \\
History = repeated current frame & 0 & 1 & 1 & 1 \\
\bottomrule
\end{tabular}
\end{table*}

\subsection{Training with Outcome-Labeled Failure Rollouts}\label{e.2.-training-with-outcome-labeled-failure-rollouts}

Because LEAF's replay dataset contains rich failure trajectories alongside successful executions, a natural question is whether naive outcome-conditioned imitation learning enables learned policies to avoid traps or recover at test time. We fine-tune \(\pi_{0.5}\) and DM0.5-MEM on combined successful and failed trajectories from the {62} {L2}--{L5} training instances ({31} tasks \(\times\) {2} training seeds), appending {Outcome:\ success} or {Outcome:\ failure} to the task instruction during training and conditioning on {Outcome:\ success} at inference. Total optimization steps (or training sample exposure) are matched to the success-only baselines in \hyperref[tab:main]{Table~\ref*{tab:main}}.

\newpage
\textbf{\hyperref[tab:s9]{Table S8}} reports first-attempt \(A_1\) and three-attempt \(A_3\) solved counts on 31 held-out and 62 training instances with disjoint random seeds. Under success-only fine-tuning, training performance moderately exceeds or closely tracks held-out performance. DM0.5-MEM reaches 17.7\% \(A_1\) and 22.6\% \(A_3\) on training seeds, versus 16.1\% and 19.4\% on held-out seeds. After 50k steps, \(\pi_{0.5}\) reaches 9.7\% versus 6.5\% on \(A_1\), and 21.0\% versus 22.6\% on \(A_3\), for training and held-out seeds, respectively.

Training-instance evaluation measures closed-loop execution from familiar initial conditions. In the failure cases we inspected, small execution deviations were followed by unsuccessful recovery. Such deviations can move the policy into states poorly covered by the demonstrations, even when the initial scene was seen during training. These observations suggest limited recovery from execution-induced distribution shift in the evaluated policies.

With a fixed training budget, adding outcome-labeled failures reduces three-attempt success on both splits: held-out \(A_3\) drops from {5/31} to {3/31} at {30k} steps and from {7/31} to {2/31} at {50k} steps for \(\pi_{0.5}\), and from {6/31} to {1/31} for DM0.5-MEM, with corresponding training declines from {14/62} to {5/62}, {13/62} to {10/62}, and {14/62} to {6/62}. At a fixed optimization budget, failed rollouts reduce exposure to successful demonstrations. Outcome-tag conditioning does not improve recovery in these configurations, although these results do not isolate why performance decreases. Together with the inspected failures, these results suggest limited learning from failures in the tested fine-tuning setup.

\subsection{Role of Within-Attempt Temporal Context at Inference}\label{e.3.-role-of-within-attempt-temporal-context-at-inference}

DM0.5-MEM uses up to {32} earlier external-camera frames sampled at {1\ Hz} within the current attempt. We evaluate the success-only checkpoint from \hyperref[tab:main]{Table~\ref*{tab:main}} with two controls: removing historical frames, or filling all {32} history slots with the current frame. In \textbf{\hyperref[tab:s9]{Table S8}}, held-out \(A_3\) falls from {6/31} with intact history to {3/31} with history removed and {1/31} with repeated current frames. Held-out \(A_1\) falls from {5/31} to {0/31} in both cases. Within-attempt visual history thus supports multi-stage execution, while clearing the buffer at resets prevents cross-attempt adaptation without explicit episodic memory.

\clearpage
\twocolumn[{\begin{minipage}{\textwidth}\captionsetup{hypcap=false}
\centering
\captionof{table}{\textbf{Composition of the released LeRobot replay dataset.} Breakdown by primary failure category ({L1}--{L6}). One step corresponds to one {20\ Hz} low-level control frame ({89.6} simulated hours total).}\label{tab:s10}
\small
\setlength{\tabcolsep}{4pt}
\renewcommand{\arraystretch}{1.12}
\begin{tabular}{@{}
  >{\raggedright\arraybackslash}p{(\linewidth - 8\tabcolsep) * \real{0.3600}}
  >{\centering\arraybackslash}p{(\linewidth - 8\tabcolsep) * \real{0.1600}}
  >{\centering\arraybackslash}p{(\linewidth - 8\tabcolsep) * \real{0.1800}}
  >{\centering\arraybackslash}p{(\linewidth - 8\tabcolsep) * \real{0.1500}}
  >{\centering\arraybackslash}p{(\linewidth - 8\tabcolsep) * \real{0.1500}}@{}}
\toprule\noalign{}
\begin{minipage}[b]{\linewidth}\raggedright
Failure Category
\end{minipage} & \begin{minipage}[b]{\linewidth}\centering
Trajectories
\end{minipage} & \begin{minipage}[b]{\linewidth}\centering
Control Steps (M)
\end{minipage} & \begin{minipage}[b]{\linewidth}\centering
Successful
\end{minipage} & \begin{minipage}[b]{\linewidth}\centering
Failed
\end{minipage} \\
\midrule\noalign{}
L1 (Execution offset) & 173 & 0.14 & 98 & 75 \\
L2 (State restoration) & 543 & 1.11 & 240 & 303 \\
L3 (Infeasible strategy) & 265 & 0.36 & 159 & 106 \\
L4 (Misjudged geometry) & 661 & 1.19 & 140 & 521 \\
L5 (Misjudged physics) & 1,341 & 2.00 & 367 & 974 \\
L6 (Chained long-horizon) & 320 & 1.65 & 39 & 281 \\
\textbf{Total} & \textbf{3,303} & \textbf{6.45} & \textbf{1,043} & \textbf{2,260} \\
\bottomrule
\end{tabular}
\end{minipage}\vspace{12pt}}]
\section{Replay Dataset Documentation}\label{f.-replay-dataset-documentation}

We release the evaluation replay dataset in LeRobot format \cite{cadene2024lerobot}. Across evaluated models, harnesses, and failure categories, its {3,303} trajectories include {1,043} successes and {2,260} failures, spanning {6.45M} control steps at {20\ Hz}, or {89.6} simulated hours. \textbf{\hyperref[tab:s10]{Table S9}} gives the breakdown by failure category. Of these trajectories, {2,172} come from history-retaining multi-attempt runs, {640} from memory-free retries, and {491} from single-attempt diagnostics.

\textbf{Frame and trajectory structure.} Each {20\ Hz} control frame records:

\begin{itemize}
\item
  synchronized {256\ ×\ 256} RGB images from the external {agentview} and wrist cameras;
\newpage
\item
  an {8-D} proprioceptive state vector containing end-effector Cartesian position, axis-angle orientation, and left/right finger positions;
\item
  the executed {7-D} low-level action: normalized end-effector pose delta and gripper command generated by the {move\_eef} controller; and
\item
  step, command, and attempt indices.
\end{itemize}

{\widowpenalty=10000 Each attempt on the {34} resettable tasks forms one trajectory. For the six non-resetting {L2}\textsuperscript{$\dagger$} tasks in \hyperref[tab:s1]{Table S1}, an episode forms one continuous trajectory, labeled successful if its final attempt meets all goal predicates. Each trajectory also stores task instructions, privileged trap metadata, real-time evaluator trap/subgoal event logs, agent-visible tool replies, and public reasoning messages exposed by the model's harness and aligned to command boundaries.\par}

\FloatBarrier